\documentclass[preprint,review]{elsarticle}

\usepackage{geometry}
\usepackage[T1]{fontenc}
\usepackage{caption}
\usepackage{subcaption}
\usepackage{xcolor}
\usepackage[linesnumbered,ruled,vlined]{algorithm2e}
\usepackage{comment}
\usepackage{array}
\usepackage{amsmath}
\usepackage{bbm}
\usepackage{tabularx}
\usepackage{calc}
\usepackage{textcomp}
\usepackage{stfloats}
\usepackage{url}
\usepackage{verbatim}
\usepackage{graphicx}
\def\BibTeX{{\rm B\kern-.05em{\sc i\kern-.025em b}\kern-.08em
    T\kern-.1667em\lower.7ex\hbox{E}\kern-.125emX}}
\usepackage{balance}
\usepackage{amssymb}
\usepackage{anyfontsize}
\usepackage{lipsum}
\usepackage{times}
\usepackage{epsfig}
\usepackage{bm}
\usepackage{wrapfig}
\usepackage{bbding}
\usepackage{ragged2e}

\usepackage{setspace}
\usepackage[percent]{overpic}

\DeclareMathOperator*{\argmax}{arg\,max}

\newcommand{\mycolor}{black}

\usepackage{tikz}
\usetikzlibrary{fit,calc}

\colorlet{neg}{red!40}
\colorlet{pos}{cyan!60}

\journal{Pattern Recognition}

\usepackage{booktabs}

\usepackage{dsfont}
\usepackage{xcolor}
\usepackage{multirow}
\usepackage[normalem]{ulem}
\useunder{\uline}{\ul}{}

\DeclareMathAlphabet\mathbfcal{OMS}{cmsy}{b}{n}
\usepackage[symbol]{footmisc}

\title{Inter-Domain Mixup for Semi-Supervised Domain Adaptation}

\author[aff1,aff2]{Jichang Li}
\author[aff1]{Guanbin Li\corref{cor2}}
\author[aff2]{Yizhou Yu}
\cortext[cor2]{Corresponding author}

\affiliation[aff1]{
   organization={School of Computer Science and Engineering, Research Institute of Sun Yat-sen University in Shenzhen, Sun Yat-sen University},
   city={Guangzhou},
   country={China}
}

\affiliation[aff2]{
   organization={Department of Computer Science, The University of Hong Kong},
   city={Hong Kong},
   country={Hong Kong}
}

\ead{csjcli@connect.hku.hk}
\ead{liguanbin@mail.sysu.edu.cn}
\ead{yizhouy@acm.org}

\begin{document}
\begin{frontmatter}

\renewcommand*{\thefootnote}{\fnsymbol{footnote}}

\begin{abstract}
Semi-supervised domain adaptation (SSDA) aims to bridge source and target domain distributions, with a small number of target labels available, achieving better classification performance than unsupervised domain adaptation (UDA). However, existing SSDA work fails to make full use of label information from both source and target domains for feature alignment across domains, resulting in label mismatch in the label space during model testing. This paper presents a novel SSDA approach, Inter-domain Mixup with Neighborhood Expansion (IDMNE), to tackle this issue. Firstly, we introduce a cross-domain feature alignment strategy, Inter-domain Mixup, that incorporates label information into model adaptation. Specifically, we employ sample-level and manifold-level data mixing to generate compatible training samples. These newly established samples, combined with reliable and actual label information, display diversity and compatibility across domains, while such extra supervision thus facilitates cross-domain feature alignment and mitigates label mismatch. Additionally, we utilize Neighborhood Expansion to leverage high-confidence pseudo-labeled samples in the target domain, diversifying the label information of the target domain and thereby further increasing the performance of the adaptation model. Accordingly, the proposed approach outperforms existing state-of-the-art methods, achieving significant accuracy improvements on popular SSDA benchmarks, including DomainNet, Office-Home, and Office-31.
\end{abstract}


\end{frontmatter}

\renewcommand*{\thefootnote}{\arabic{footnote}}

\newcommand{\currprop}{\textwidth}


\setlength\tabcolsep{4.5pt} 

\section{Introduction}
\label{Section:Introduction}

Domain adaptation (DA) aims to first train an adaptation model from label-rich datasets (a.k.a source domain) and then transfer the learned knowledge to a new but label-scarce dataset (a.k.a target domain) of different distribution so as to avoid relying largely on the human-annotated in-distribution data. Nowadays, application scenarios of DA include semantic segmentation~\cite{PR-Seg}, object detection~\cite{PR-OjD}, person re-identification~\cite{PR-PID}, and so on. Unfortunately, a direct application of such models trained on the source domain dataset to the target domain would cause severe performance degradation due to different data distributions in these two domains. The commonly defined DA setting, unsupervised domain adaptation (UDA), witnesses great progress in the reduction of domain shift~\cite{ganin2015unsupervised}.  Nevertheless, compared with UDA, semi-supervised domain adaptation (SSDA), where a small number of target labels are available, achieves much better performance on the target domain. This is because supervision on a few labeled target domain samples, as well as a large number of labeled source domain samples, is already capable of bridging partial distribution discrepancies across domains~\cite{saito2019semi, li2021cross}.

\begin{figure}[t]
\centering
\includegraphics[width=9.0cm,height=4.2cm]{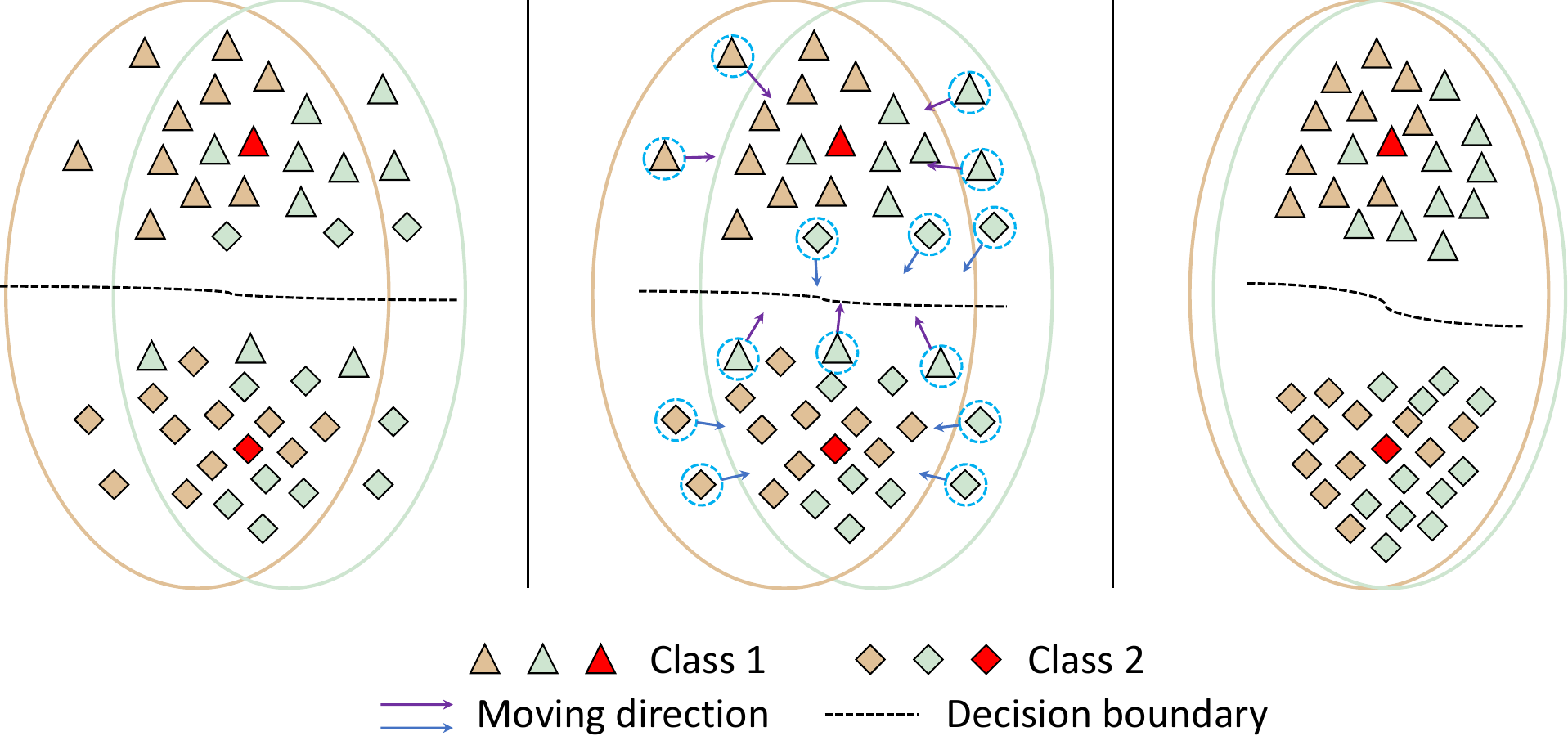}
\caption{A conceptual description of our basic idea. Sample points in brown, green, and red represent source domain data, target domain data, and class prototypes, respectively. Arrows in purple indicate that sample points move towards the prototype of Class 1, while blue arrows illustrate that the prototype of Class 2 attracts samples from the corresponding class towards itself. {\bf Left:} Previous label-free strategies to enforce domain-level feature alignment fail to generate discriminative target features, thereby giving rise to cross-domain label mismatch in label space. {\bf Middle:} Our approach incorporates label information into adaptation, and thus, the model can align class-wise sample features from both domains with the aid of their class labels. {\bf Right:} The proposed approach enables the model to produce domain-invariant and discriminative features and thus enhance the performance of the model.
}
\label{Figure-Idea}
\end{figure}

Previous approaches for the SSDA problem extract domain-invariant features mainly by minimizing cross-domain discrepancy measures~\cite{kim2020attract}, relying on image style transfer~\cite{luo2021relaxed} or adversarial training~\cite{saito2019semi, qin2021contradictory, li2021cross}. They can achieve domain alignment so that, in theory, or in hypothesis, the adapted classifier is prone to obtain good classification performance in the target domain. However, these strategies to enforce domain-level feature alignment may fail to generate discriminative target features for two main reasons.  First, the source domain has a much larger number of labeled samples than the target domain when we perform supervised learning using labeled samples across domains. Thus, features of labeled target domain samples do not have the same level of diversity as those of source domain samples~\cite{li2020attribute, li2021cross}, impairing their discriminability. In addition, 
\textcolor{\mycolor}{the majority of previous strategies performed to cross-domain feature alignment}
neglect label information from either the source or target domain for adaptation and the model consequently cannot align sample features from both domains according to their class labels~\cite{kang2019contrastive}. As a result, as shown in Figure~\ref{Figure-Idea}, domain-invariant yet non-discriminative features generated from the model are used to align different categories, thereby giving rise to cross-domain label mismatch in the label space~\cite{deng2019cluster}. To alleviate this issue, existing label-free feature alignment strategies have to impose more crafted constraints on the target domain~\cite{li2021cross, kim2020attract, YangLuyu2020DCwT}; nevertheless, unsupervised learning based regularization is not necessarily beneficial to label mismatch rectification, but on the contrary, may make it worse.  For example, entropy minimization in~\cite{saito2019semi}, or self-training in~\cite{li2021cross}, seeks to learn knowledge from the model predictions themselves, which is risky due to label noise accumulation~\cite{chen2019progressive} and poor model calibration~\cite{guo2017calibration}.  
\textcolor{\mycolor}{
Hence, recent advances, such as \cite{luo2021relaxed, R1d}, have proposed label-aware alignment strategies for cross-domain feature distributions. which considers the ingratiation of label information into adaptation, thereby effectively mitigating the aforementioned issues. However, these alignment strategies overlook the potential to delve deeper into the provided labeled source and target domains to excavate more definitive and authentic supervised label information. Such exploration of a broader cross-domain search scope could better bridge distribution discrepancies between domains, thereby encouraging the model to generate domain-invariant yet discriminative features on the target domain.
}

This paper  proposes a novel label-aware cross-domain feature alignment strategy, namely Inter-domain Mixup, where label information is incorporated into model adaptation. Specifically, Inter-domain Mixup conducts sample-level data mixing and manifold-level data mixing between paired labeled samples from the source and target domains. In detail, sample-level data mixing directly mixes the source images and target images and their corresponding labels through linear combinations of existing labeled samples; on the other hand, manifold-level data mixing creates virtual samples with convex combinations of the features (outputs at the penultimate layer of the model) along with their labels from the source-target sample pairs from labeled data. Thus, many virtual training samples with reliable label information are created. These newly established samples along with their labels connect both domains and are both diverse and complementary to both domains. We, hence, perform supervised learning using these labeled virtual samples so as to mitigate the discrepancy between the source and target domains. As a result, the generated features are agnostic to the distribution discrepancies between the two domains, and meanwhile, exhibit class discriminability. 

As introduced above, Inter-domain Mixup makes full use of label information from the labeled samples to enforce label-aware feature alignment across domains. However, there is only a very small amount of labeled data in the target domain, which seriously hinders the potential of cross-domain fusion because the diversity of samples in the target domain is not reflected in data mixing. 
As the model's generalization ability continues to improve during the training process, for unlabeled samples highly correlated to labeled ones, their predicted labels may become increasingly reliable. Like label propagation for semi-supervised learning~\cite{kipf2017semi, iscen2019label}, to better leverage unlabeled samples in the target domain, we introduce Neighborhood Expansion to transfer existing label information to unlabeled samples. Specifically, we perform label propagation via pseudo labeling to progressively produce pseudo class labels for unlabeled target domain samples that have high-confidence model predictions for the corresponding classes.
In the meantime, two schemes, i.e., Self-Regularization and Pairwise Approaching, are presented to reduce the uncertainty of model predictions at unlabeled samples in the target domain, which benefits the model in producing low-entropy and high-confidence predictions. 

We call this SSDA framework {\bf I}nter-{\bf d}omain {\bf M}ixup with {\bf N}eighborhood {\bf E}xpansion ({\bf IDMNE}). In a nutshell, the main contributions of our proposed method can be summarized as follows:


\begin{itemize}
    \item Inter-domain Mixup, a novel cross-domain feature alignment strategy, is proposed to conduct sample-level and manifold-level data mixing for source-target sample pairs from labeled data, which not only facilitates cross-domain feature alignment but also alleviates label mismatch simultaneously to address cross-domain distribution discrepancies and thus achieve a considerable performance gain for model adaptation. 
    
    \item Neighborhood Expansion is proposed to leverage massive high-confidence pseudo-labeled target domain samples to diversify the label information of the target domain. Moreover, Neighborhood Expansion includes Self-Regularization and Pairwise Approaching, which reduce the uncertainty of model predictions at unlabeled target domain samples in order to make them more confident.
    
    \item Inter-domain Mixup and Neighborhood Expansion are integrated into an adaptation framework, and numerous experiments show our proposed method can outperform existing state-of-the-art approaches and achieve considerable accuracy improvement on three commonly used benchmark datasets such as \texttt{DomainNet}~\cite{PengXingchao2018MMfM}, \texttt{Office-Home}~\cite{VenkateswaraHemanth2017DHNf} and \texttt{Office-31}~\cite{saenko2010adapting}.
\end{itemize}

\section{Related Work}

\subsection{Domain Adaptation}

Deep neural networks (DNNs) do well in learning discriminative representations for input data by resorting to a considerable amount of labeled data, which is extremely expensive and time-consuming to obtain. 
Recently, a vast number of domain adaptation techniques~\cite{ZhangYang2020ACDA, PR-DA-001, PR-DA-002} attempted to design good adaptation models in order to train with the source domain data with rich labels and the target domain data with scarce labels to realize the accurate recognition of the target domain data.

In general, DA requires addressing huge domain shifts by learning to align cross-domain feature representations. 
Mainstream DA methods consider learning domain-invariant knowledge by constructing various statistical measures to represent domain discrepancy and then decreasing it, such as Maximum Mean Discrepancy (MMD)~\cite{pan2010domain} and its modified versions~\cite{long2015learning, long2017deep, zellinger2017central}. For instance, Long et al.~\cite{long2015learning} introduced a deep adaptation network to minimize MMD over multiple domain-specific feature representation layers so as to learn more transferable features. Also, JMMD proposed in~\cite{long2017deep}, inherited from conventional MMD, was employed to enforce cross-domain joint distribution alignment on domain-specific layers, while CMD used in~\cite{zellinger2017central} performed order-wise matching in higher-order feature distributions. 

Adversarial training, which is employed to confuse the discriminator in a two-player min-max manner to learn cross-domain feature alignment, is another effective way for aligning feature distributions across domains~\textcolor{\mycolor}{\cite{tang2020discriminative, li2021cross, Adversarial2020xu, ganin2015unsupervised}}. In particular, Ganin et al. in~\cite{ganin2015unsupervised} introduced a gradient reversal layer into the adaptation framework so that a classifier and a discriminator can be responsible for two tasks, namely, object classification and domain classification. The former trains the classifier over source domain data while the latter learned domain-invariant feature representations across domains by fooling the discriminator. Also, Tang et al.~\cite{tang2020discriminative} proposed discriminative adversarial learning to carry out domain alignment at both the feature level and category level. 

Furthermore, several recent work~\cite{kim2020learning, luo2020adversarial} also consider image style transfer across domains to bridge visual domain differences as DNN is sensitive to the style of image inputs as pointed out in~\cite{nam2018batch}. For instance, Kim~et al.~\cite{kim2020learning} introduced a style transfer algorithm to stylize the source domain to adapt the target domain so as to diversify the texture of synthetic images. Also, Adversarial Style Mining proposed in~\cite{luo2020adversarial} explored complex styles for an unseen target domain to enhance the adaptation performance in a data-scarce scenario. 
\textcolor{\mycolor}{What's even more, other studies~\cite{R2C1, R2C2} have focused on addressing negative transfer to improve the adaptation capability of models for the target domain. Specifically, Lu et al. in ~\cite{R2C1} proposed weighted correlation embedding learning to specifically address what should be transferred for a given task, thereby avoiding negative transfer caused by distribution outliers. In addition, Lu et al. in ~\cite{R2C2} 
 introduced guided discrimination and correlation subspace learning for cross-domain image classification, which accounts for domain-invariant, category-discriminative, and correlation-based learning of data. Last but not least, Lu et al. in~\cite{R2C3} also put forward a method called cross-domain structure learning (CDSL) for recognizing visual data in the target domain. CDSL incorporates global distribution alignment and local discriminative structure preservation to extract common underlying features between domains.}

In practice, domain adaptation has been developed for many new settings in different application scenarios, such as heterogeneous domain adaptation~\cite{WenLi2014LWAF}, open set domain adaptation~\cite{kunducvpr2020},
partial domain adaptation~\cite{8951442}, etc. In this paper, we focus on the tasks of domain adaptation related to semi-supervised learning due to its potential superiority over the commonly defined DA setting, i.e., unsupervised domain adaptation.

\subsection{Semi-supervised Domain Adaptation}

Tremendous progress has been made in cross-domain feature alignment by recent DA approaches~\cite{li2021cross, ZhangYang2020ACDA, saito2019semi} in order to reduce distribution discrepancies between both domains to some extent. 
However, commonly used DA techniques, such as decreasing cross-domain statistical discrepancy measures, domain adversarial learning, and image style transfer, etc., can only ensure that the model produces domain-invariant features, but as label information is not incorporated in constraining model training, the discriminability of generated features cannot be guaranteed, thus is of considerable significance in matching class-wise distributions~\cite{chen2019transferability, you2021learning}. Therefore, several recent DA methods~\cite{Adversarial2020xu, deng2019cluster} proposed to impose constraints on the target domain to take into consideration the class-aware information so as to alleviate label mismatch across domains, {e.g.}, assigning pseudo-labels for unlabeled samples in the target domain. 

In contrast to the above practice, Saito et al. in~\cite{saito2019semi} directly assumed that the target domain has a small fraction of ground-truth labels (typically one-shot or three-shot per class), which achieves significant performance gains through extra supervision on the target domain. However, Saito et al. in~\cite{saito2019semi} also demonstrated that a direct application of traditional UDA techniques to the SSDA problem does harm its effectiveness. Therefore, they proposed Minimax Entropy which optimizes in an adversarial scheme to minimize the distance between the class prototypes and neighboring unlabeled target domain samples so as to achieve domain alignment.  This relied heavily on a few target labels to establish correlations among samples from both domains. Similar to~\cite{saito2019semi}, Qin et al.~\cite{qin2021contradictory} also introduced an adversarial learning based method called Contradictory Structure Learning to enforce a target-classifier and a source-classifier so as to learn well-clustered target features and well-scattered source features, respectively. In this case, this method could better align the target distribution with the source distribution. Li~et al.~\cite{li2021cross} proposed a cross-domain adaptive clustering algorithm to achieve cluster-wise feature alignment across domains, in which adversarial learning was also adopted. In addition, APE proposed in~\cite{kim2020attract} attempted to adopt an attraction scheme to align global feature distributions across domains and then take on a perturbation scheme and an exploration scheme to optimize intra-domain discrepancy in the target domain. Compared to the above feature-level adaptation, Luo et al. in~\cite{luo2021relaxed} designed a Relaxed-cGAN to produce new image-label pairs with class-wise conditional semantic information for pixel-level adaptation through image transfer, where the information of source images is ignored. 

To sum up, despite the success of previous label-free cross-domain feature alignment strategies to address the SSDA tasks, most of them focus on aligning features at the domain level and ignore the label attribution of samples. In this case, the generated features cannot be aligned according to their class labels, perhaps giving rise to label mismatch in the label space. Hence, imposing crafted constraints on the target domain is crucial for these strategies. Instead, our proposed Inter-domain Mixup takes advantage of the label information of labeled samples to learn class-wise feature alignment across domains by incorporating them into adapting the model. This ensures samples of diverse classes are aligned correctly, especially on the target domain.

\subsection{Data Mixup}

Data Mixup~\cite{zhang2018mixup} refers to performing convex interpolation on a pair of labeled samples to generate augmented samples for model training. It is designed as a regularizer and augments the smoothness of learned features for supervised learning. Verma et al.~\cite{pmlr-v97-verma19a} extended it to produce more continuous hidden representations during training. Recently, Mixup has confirmed its effectiveness in the field of semi-supervised learning (SSL)~\cite{NEURIPS2019_1cd138d0, remixmatch}. For example, Mixmatch augmented labeled and unlabeled samples for training to smooth the model's manifolds~\cite{NEURIPS2019_1cd138d0}. Also, Berthelot~et al.~\cite{remixmatch} enhanced Mixmatch and then proposed ReMixMatch for training the SSL model to achieve a better classification performance. As well, Wang et al.~\cite{Wang_2020_CVPR} used Mixup to deal with the problem of semi-supervised 3D Medical Image Detection and obtained a substantial improvement in performance by mixing medical images at the image level and object level. Afterwards, several studies~\cite{virtualmixup, dualmixup, na2021fixbi} found the potential of applying Mixup in domain adaptation. For instance, Wu~et al.~\cite{dualmixup} proposed two mixup regularizers at the category and domain levels to instruct the classifier to enforce the consistency of in-between sample predictions and to enrich feature-space intrinsic structures. Also, Na~et al. presented in~\cite{na2021fixbi} augmented diverse intermediate domains between source-target sample pairs using a fixed ratio-based mixup regularizer, which successfully bridged domain spaces so as to alleviate domain discrepancy.

In this paper, we use Data Mixup to establish Inter-domain Mixup, a novel strategy to perform cross-domain feature alignment for the SSDA task.  The most similar work to our method may be~\cite{LiXingjian2020XETL, chen2021semiseg, R3C1}, but there are obvious differences. Firstly, our model behaves linearly within source-target labeled sample pairs whose labels are available randomly. Instead,~\cite{LiXingjian2020XETL} requires pairing each target class with a unique and dedicated source class, i.e., one-to-one pairing from the target to source classes. 
\textcolor{\mycolor}{
In addition, the limitation by the absence of target labels makes these works namely\cite{R3C1, LiXingjian2020XETL, chen2021semiseg} an unsupervised Mixup approach that lacks actual but diversified supervised label information. This unsupervised scheme might inadvertently introduce noisy label information, leading to more serious label mismatch during domain alignment. Moreover, the features generated in \cite{R3C1, LiXingjian2020XETL, chen2021semiseg} lack associated real label information, thus limiting the classifier's discriminative efficacy. In contrast, the actual but virtual sample-label pairs constructed by our proposed method are fed into the classifier for training, thereby enhancing the classifier's discriminatory capacity.}
Also,~\cite{chen2021semiseg} has been proposed to effectively address the task of semantic segmentation, whereas our approach is focused on enhancing the performance of image classification.
Most importantly,~\cite{LiXingjian2020XETL, chen2021semiseg} only enforce data mixing at the sample level but ours conducts convex interpolation at both sample and manifold levels. Our proposed method encourages the model to explore more cross-domain searching ranges so as to better bridge distribution discrepancies between domains.
\textcolor{\mycolor}{
Although~\cite{R3C1} takes feature-level Mixup into account, pursues the approximation of mixed features to mixup input features through the MSE loss. This scheme yields a narrower cross-domain search scope, thereby compromising the effective bridging of distribution discrepancies across domains.
}

\begin{figure*}[t]
    \centering
    \includegraphics[width=10.0cm,height=4.64cm]{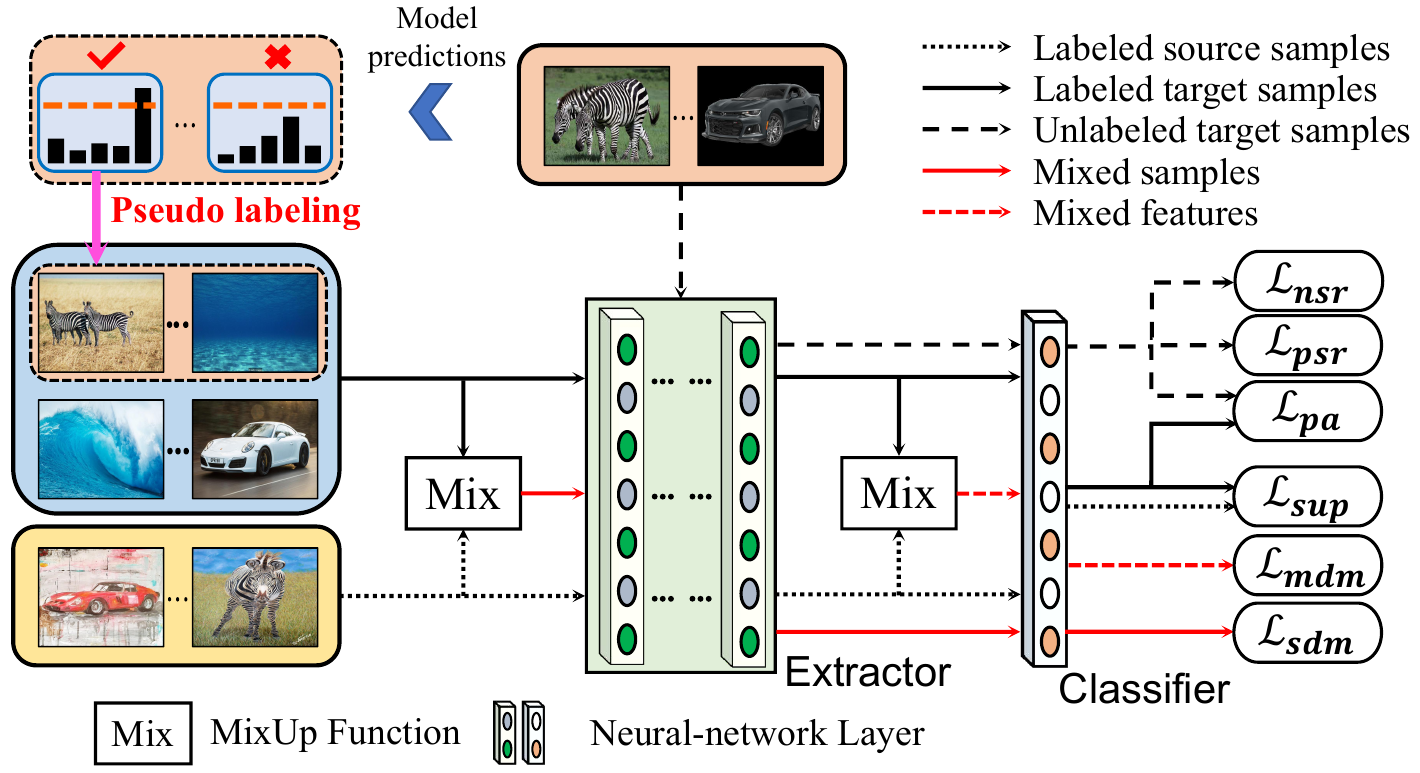}
    \caption{An overview of our proposed Inter-domain Mixup with Neighborhood Expansion for semi-supervised domain adaptation. We use arrows with different line styles to represent data flow, where the black arrow denotes labeled source or target domain data, and the red arrow indicates mixed sample or mixed feature. Our model includes an extractor for feature generation and a classifier for object classification. Also, we train our model with six loss terms in which $\mathcal{L}_{\bm{sup}}$ is for supervision over labeled data from both domains; $\mathcal{L}_{\bm{sdm}}$ and $\mathcal{L}_{\bm{mdm}}$ are for Inter-domain Mixup to perform cross-domain class-wise feature alignment; and the remaining $\mathcal{L}_{\bm{psr}}$, $\mathcal{L}_{\bm{nsr}}$ and $\mathcal{L}_{\bm{pa}}$ are for Neighborhood Expansion to make unlabeled target domain data more confident. To further leverage unlabeled samples in the target domain, we also employ pseudo labeling to assign pseudo-labels to unlabeled target domain samples with high probability scores and merge the selected pseudo-labeled target domain samples into the labeled target domain set.}
    \label{Figure-Overview}
\end{figure*}

\section{The Proposed Method}

In this section, we first present the problem formulation and notations in the context of semi-supervised domain adaptation (SSDA), and then elaborate on our proposed method, i.e., Inter-domain Mixup with Neighborhood Expansion (IDMNE). An overview of the proposed method is illustrated in Figure~\ref{Figure-Overview}.

\subsection{Problem Formulation and Notation}
In the context of SSDA, we are given two sets of labeled samples in the source and target domains respectively, and a set of unlabeled samples in the target domain. They can be denoted as $\mathcal{D}_{s}={\{(x_i^s,y_i^s)\}}_{i=1}^{N_s}$, $\mathcal{D}_{l}={\{(x_i^{l},y_i^{l})\}}_{i=1}^{N_{l}}$ and $\mathcal{D}_{u}={\{(x_i^{u},)\}}_{i=1}^{N_{u}}$, respectively, where the size of $\mathcal{D}_{l}$, i.e., $N_{l}$, is much smaller than $N_s$ and $N_{u}$. Specifically, each sample $x_i^s$ ($x_i^{l}$) in the labeled sample set $\mathcal{D}_{s}$ ($\mathcal{D}_{l}$) is accompanied with a given label indexed by $y_i^s$ ($y_i^{l}$). However, for an unlabeled target domain sample $x_i^u$ from $\mathcal{D}_{u}$, we have no access to its associated label. Provided with the aforementioned information, the goal of this work is to learn an adaptation model that enables accurate classification of target domain samples during the testing phase.

Recent SSDA work~\cite{saito2019semi, kim2020attract, li2021cross, YangLuyu2020DCwT, luo2021relaxed, berthelotadamatch, SSDA-MCL} has proved that a prototypical classifier is beneficial to feature alignment across domains. Following such studies, we also construct a model with parameters $\theta$, which consists of a feature extractor $\mathcal{F}$ and a prototypical classifier $\mathcal{G}$. The extractor is a deep neural network followed by an $\ell_{2}$ normalization layer, while the classifier comprises an unbiased linear layer. According to~\cite{saito2019semi, kim2020attract,li2021cross}, the weights of such a prototypical classifier can be represented as class prototypes,~i.e., $\mathbf{W}=[\mathbf{w}_1, \cdots,\mathbf{w}_k, \cdots, \mathbf{w}_K]$, where $k = 1, 2, ..., K$ indicates the class index. In this case, a normalized feature $\frac{\mathcal{F}(x)}{\lVert \mathcal{F}(x) \rVert}$ of a data point $x$ is fed into the classifier, meaning that the feature has been mapped into a spherical feature space~\cite{saito2019semi, kim2020attract}. Then, the distances between this point and the class prototypes $\{\mathbf{w}_k\}_{k=1}^{K}$ can be represented as the probabilistic prediction outputs of the classifier,~i.e.,
\begin{equation}
    \mathbf{p}=p(x)=\sigma(\mathcal{G}(\mathcal{F}(x)))=\sigma(\frac{1}{T}\frac{{\mathbf{W}}^{\top}\mathcal{F}(x)}{\lVert \mathcal{F}(x) \rVert}),
\end{equation}
where $\sigma(\cdot)$ represents a softmax function and $T$ indicates a temperature parameter. Larger $p_k(x)$, where $p_k(x)$ is the $k$-th component of $p(x)$, means a higher correlation between the data point $x$ and the prototype corresponding to Class $k$, i.e., $\mathbf{w}_k$. To build a good model for adaptation, we have to minimize the distances between the class prototypes and corresponding samples from both source and target domains, indirectly strengthening correlations between both domains and thus achieving cross-domain feature alignment.

In this paper, we perform supervised model training over all labeled samples across domains to address cross-domain distribution discrepancies using a standard cross-entropy loss as follows,
\begin{equation}
{\mathcal{L}}_{\bm{sup}}(\theta;{\mathcal{D}_{s}, \mathcal{D}_{l}})=-\frac{1}{N_s+N_l}\sum_{(x_i, y_i)\in{\mathcal{D}_{s}\cup\mathcal{D}_{l}}}p_y({y}_i)\log(p(x_i)),
\end{equation}
where $p_y(\cdot)$ is a function to create a one-hot label probability vector corresponding to the index of a class label.

\begin{figure*}[t]
\centering
\includegraphics[width=9.0cm,height=2.45cm]{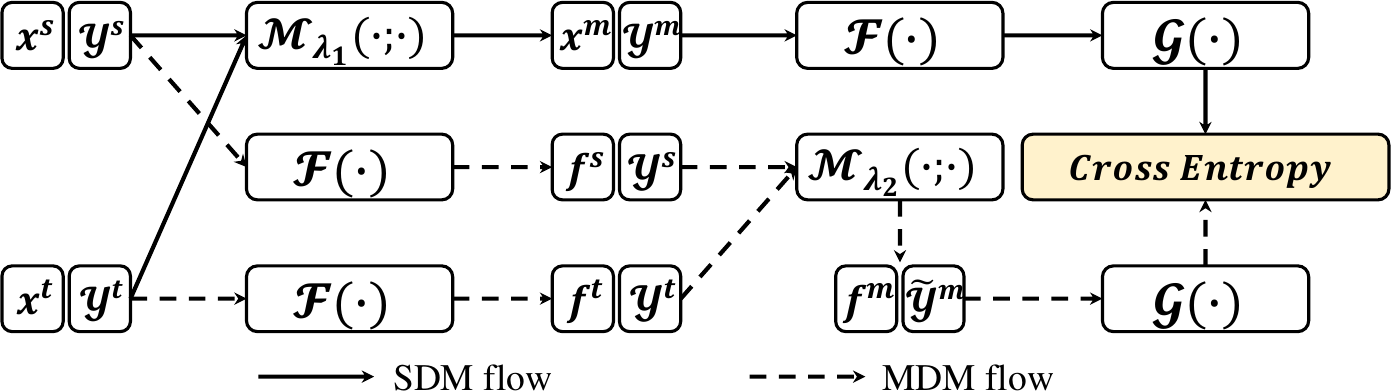}
    \caption{A flow diagram of Inter-domain Mixup. A solid line represents a flow of sample-level data mixing (SDM), while a dashed line indicates a flow of manifold-level data mixing (MDM). $\mathcal{M}_{\lambda_{1}}(\cdot;\cdot)$ and $\mathcal{M}_{\lambda_{2}}(\cdot;\cdot)$ are two mixup functions where $\lambda_{1}$ and $\lambda_{2}$ are two different mixup ratios. For the SDM flow, we obtain a mixup sample $(x^m, y^m)$ by mixing a source-target sample pair containing a labeled source domain sample $(x^s, y^s)$ and a labeled target domain sample $(x^t, y^t)$ through a linearly convex interpolation. For the MDM flow, two feature representations $f^s$ and $f^t$ along with their original labels $y^s$ and $y^t$ are mixed to generate an augmented feature $f^m$ and its associated label $\tilde{y}^m$. Afterwards, extra supervision over these two types of mixup points, i.e., $(x^m, y^m)$ and $(f^m, \tilde{y}^m)$, is performed via a standard cross-entropy loss function.}
    \label{Figure-Diagram-IDM}
\end{figure*}

\subsection{Inter-Domain Mixup}
\label{Subsection:IDM}

Similar to other DA tasks~\cite{saito2019semi, kim2020attract, li2021cross, LiLusi2020DAfP, kundu2020cvpr}, we need to consider data from the source and target domains in the context of probability distributions ${\mathcal{P}}_s$ and ${\mathcal{P}}_t$, where ${\mathcal{P}}_{s}\neq{\mathcal{P}}_{t}$. Therefore, our adaptation model needs to reduce domain shifts in order to align feature distributions across domains. 

As partial label information is available in the target domain, we propose to make full use of the label information in the target domain to achieve domain adaptation. Specifically, we propose a novel cross-domain feature alignment strategy called Inter-Domain Mixup, where source domain samples and target domain samples, as well as their affiliated labels, are integrated into the feature alignment process with Mixup operations. It acts as data augmentation and can produce additional reliably labeled samples for model training. Also, in the inter-domain mixup operation, each generated sample can be regarded as a bridge between the source and target domains, playing an indispensable role in closing the gaps between the two domains. Hence, extra supervision from these intermediate and complementary data not only prevents the model from overfitting the source domain dataset but also improves feature discriminability. Our proposed Inter-Domain Mixup conducts sample-level data mixing ({\bf SDM}) and manifold-level data mixing ({\bf MDM}) on source-target labeled sample pairs. A flow diagram of Inter-domain Mixup is given in Figure~\ref{Figure-Diagram-IDM}.

{\bf SDM} is achieved with vanilla Mixup~\cite{zhang2018mixup} to produce mixed image $x^m$ and its corresponding mixed label $y^m$ through a convex interpolation of a source-target sample pair as follows:
\begin{equation}
    \begin{split}
        {x^m}=&\mathcal{M}_{\lambda_{1}}(x^s, x^t)={{\lambda}_{1}}{x^s}+({1-{\lambda}_{1}}){x^t}, \\
        {y^m}=\mathcal{M}&_{\lambda_{1}}(y^s, y^t)={{\lambda}_{1}}{p_y(y^s)}+({1-{\lambda}_{1}}){p_y(y^t)},
    \end{split}
\end{equation}
where ${\lambda}_1 \sim Beta(\alpha)$ is a mixup ratio and $\alpha$ is a constant scalar of the beta distribution. As well, ($x^s, y^s$) and ($x^t, y^t$) are labeled samples from ${D}_{s}$ and ${D}_{l}$ respectively. Note that ${y^m}$ here is not a class index but a label probability vector.

Meanwhile, $(x^s, y^s)$ and $(x^t, y^t)$ are fed into the feature extractor and the feature representations ${f^s}=\mathcal{F}(x^s)$ and ${f^t}=\mathcal{F}(x^t)$ are obtained. According to Manifold Mixup used in~\cite{pmlr-v97-verma19a}, we carry out {\bf MDM} that mixes the source feature ${f^s}$ and the target feature ${f^t}$ in the latent feature space by means of a convex combination:
\begin{equation}
    \begin{split}
        {f^m}=&\mathcal{M}_{\lambda_{2}}(f^s, f^t)={{\lambda}_{2}}{f^s}+({1-{\lambda}_{2}}){f^t}, \\
        {\tilde{y}^m}=\mathcal{M}&_{\lambda_{2}}(y^s, y^t)={{\lambda}_{2}}{p_y(y^s)}+({1-{\lambda}_{2}}){p_y(y^t)},
    \end{split}
\end{equation}
where ${\lambda}_2 \sim Beta(\alpha)$ is another mixup ratio.

Compared with existing labeled samples from both domains, the mixed samples or features associated with mixed labels obtained by SDM and MDM enrich label information and create a denser data distribution in-between the two domains. Finally, we impose extra supervision on the mixed samples using the following loss,
\begin{equation}
    \label{loss:IDM}
    \begin{split}
        {\mathcal{L}}_{\bm{IDM}}(\theta;{\mathcal{D}_{s}, \mathcal{D}_{l}})={\mathcal{L}}_{\bm{sdm}}(\theta;{\mathcal{D}_{s}, \mathcal{D}_{l}}) + {\mathcal{L}}_{\bm{mdm}}(\theta;{\mathcal{D}_{s}, \mathcal{D}_{l}}),
    \end{split}
\end{equation}
\begin{equation}
    \label{Equation:SDM}
    \begin{split}
        {\mathcal{L}}_{\bm{sdm}}(\theta;{\mathcal{D}_{s}, \mathcal{D}_{l}})=-\frac{1}{N_{pair}}\sum_{i=1}^{N_{pair}}{y^m_i \log{(p(x^m_i))}},
    \end{split}
\end{equation}
\begin{equation}
    \label{Equation:MDM}
    \begin{split}
        {\mathcal{L}}_{\bm{mdm}}(\theta;&{\mathcal{D}_{s}, \mathcal{D}_{l}})=-\frac{1}{N_{pair}}\sum_{i=1}^{N_{pair}}{\tilde{y}^m_i \log{(\sigma(\mathcal{G}(f^m_i)}))},
    \end{split}
\end{equation}
where $N_{pair}$ indicates the number of source-target sample pairs constructed by one-to-one pairing from labeled samples in ${\mathcal{D}_{s}}$ and ${\mathcal{D}_{l}}$.
 
As discussed in Section~\ref{Subsection:Analysis}, our mixup regularizer improves model calibration, thereby increasing the accuracy of pseudo-labels of unlabeled target domain samples provided by Neighborhood Expansion. 
In addition, Mixup also has the effect of denoising, attenuating the negative impact of incorrect pseudo-labels during training.
In general, we generate augmented samples by mixing labeled source domain samples with originally labeled target domain samples or pseudo-labeled target domain samples. As a result, Inter-domain Mixup ensures that at least a portion of the label information for such an augmented sample is reliable. 

\begin{figure}[t]
    \centering
    \begin{subfigure}{0.32\textwidth}
        \includegraphics[width=\linewidth]{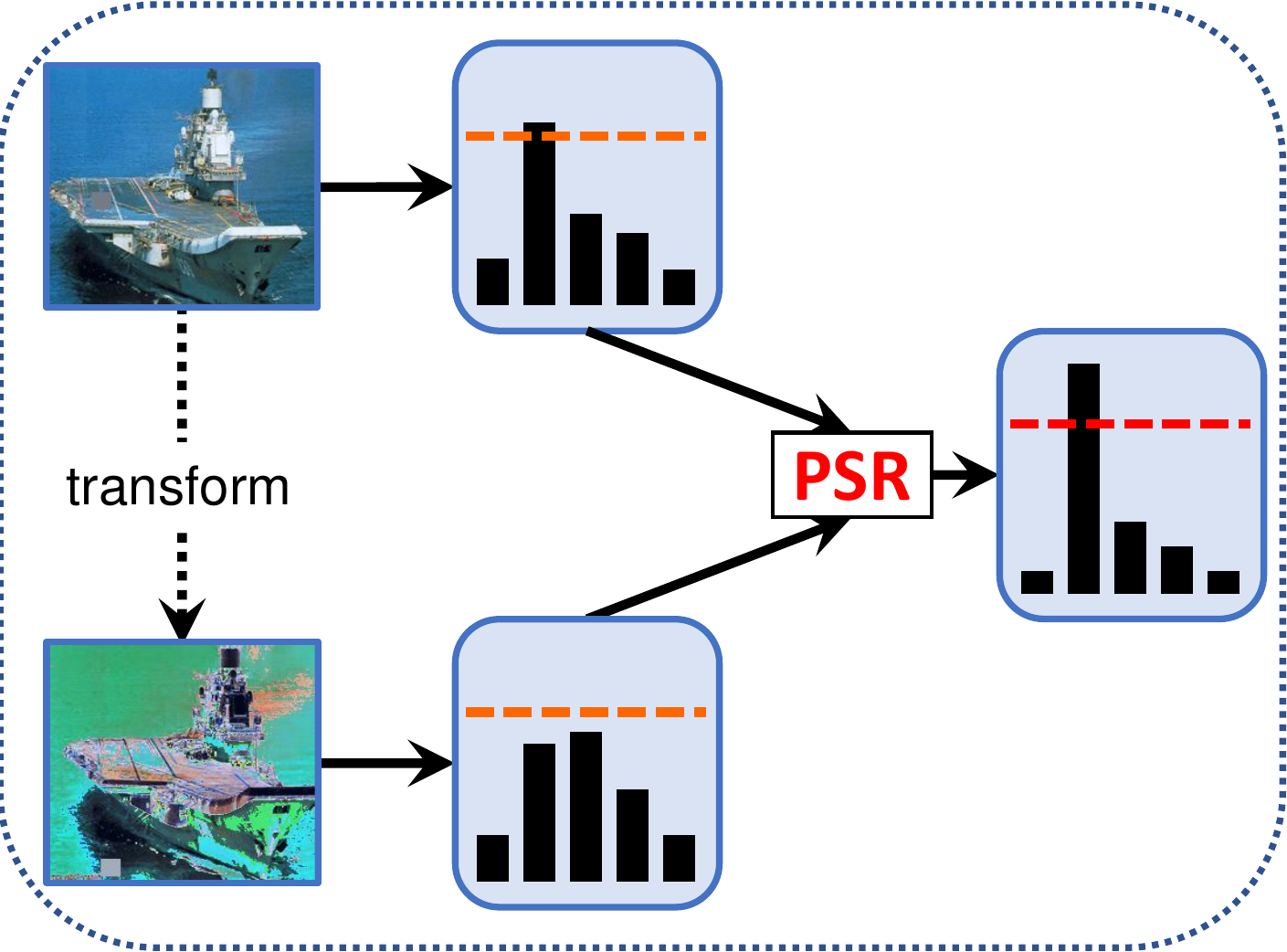}
        \caption{PSR}
    \end{subfigure}
    \hfill
    \begin{subfigure}{0.32\textwidth}
        \includegraphics[width=\linewidth]{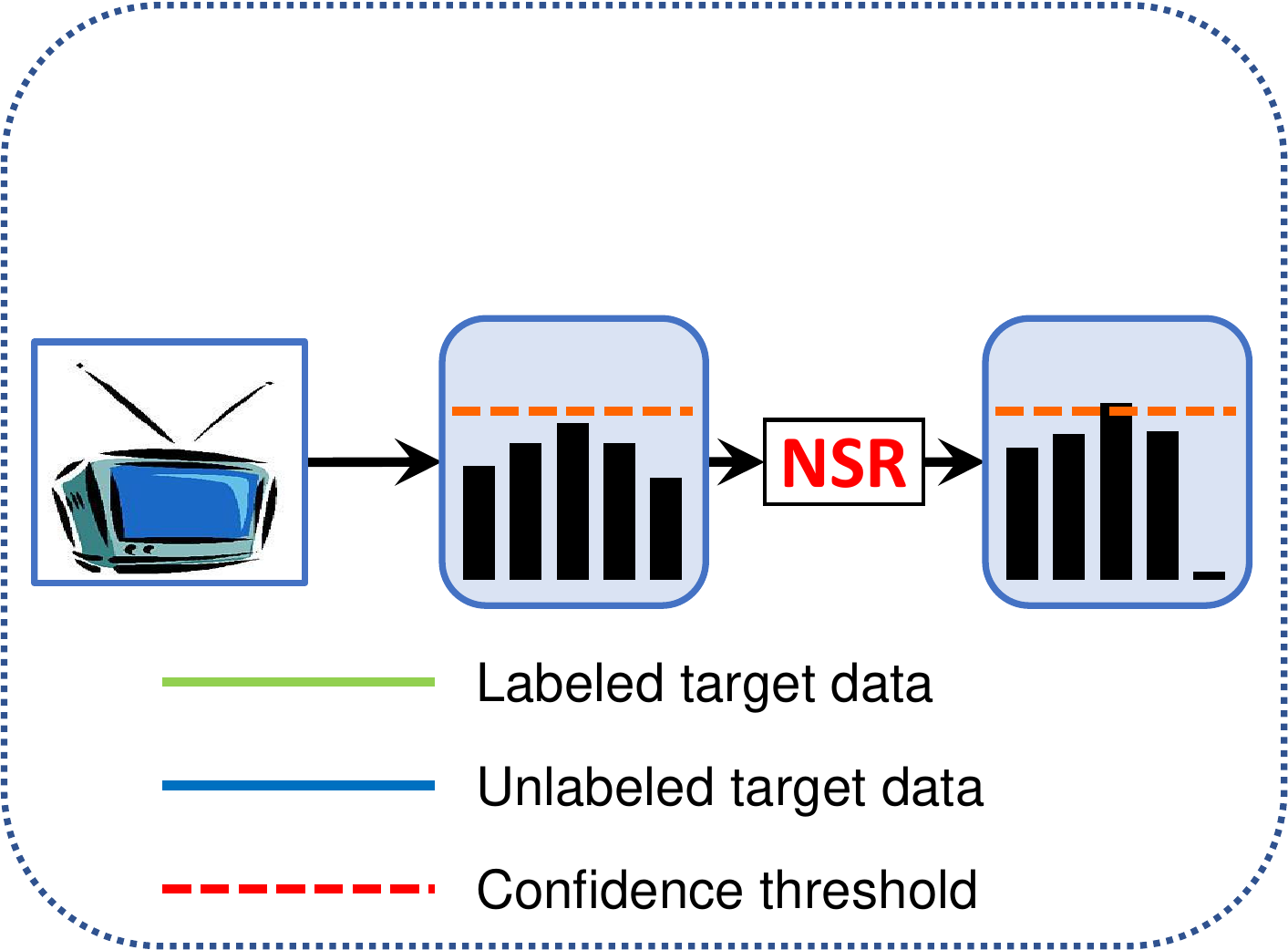}
        \caption{NSR}
    \end{subfigure}
    \hfill
    \begin{subfigure}{0.32\textwidth}
        \includegraphics[width=\linewidth]{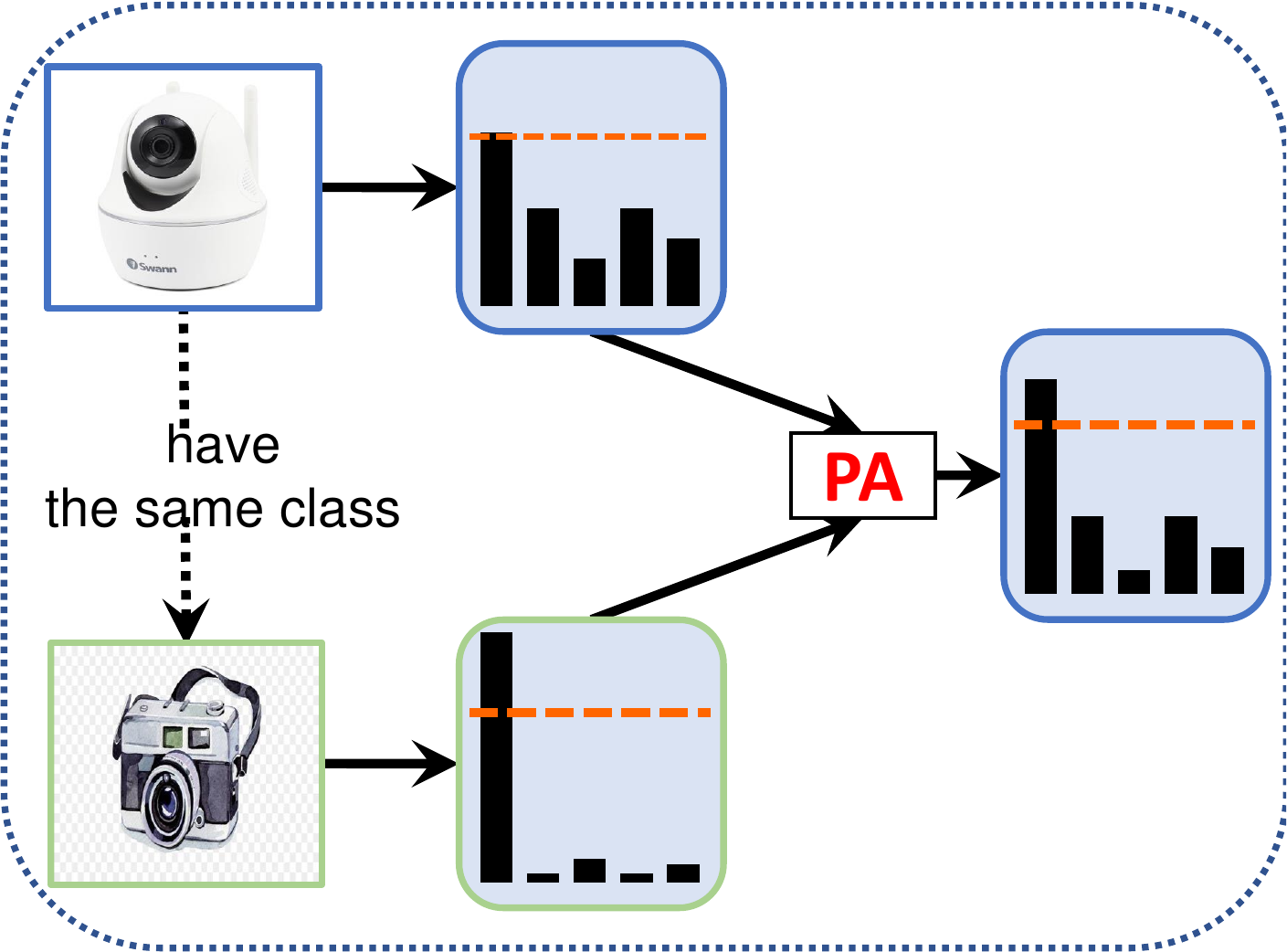}
        \caption{PA}
    \end{subfigure}
    \caption{Illustrations of three schemes applied in Neighborhood Expansion to encourage low-entropy and high-confidence predictions for unlabeled target domain samples. (a) Positive self-regularization learning (PSR) introduces self-training to augment the model's robustness. (b) Negative self-regularization learning (NSR) is to raise the predictive probabilities of each class except for the class corresponding to the lowest predicted probability scores. (c) Pairwise Approaching (PA) aims to drive high-confidence unlabeled target domain samples towards labeled data of the same class in the target domain. Note that PSR and PA handle samples from $\mathcal{D}_{u}$ whose confidence scores of their predicted class labels are above the confidence threshold $\tau$, while NSR is of importance for unlabeled target domain samples with confidence scores lower than $\tau$.}
    \label{Figure-NE}
\end{figure}

\subsection{Neighborhood Expansion}
Inter-domain Mixup benefits the model by creating a large number of valuable labeled samples. However, only a limited portion of target labels are available, i.e., one-shot or three-shot per class, which significantly restrains the diversity of mixup samples and the representativeness of the dataset. Several recent work~\cite{kim2020attract, li2021cross, YangLuyu2020DCwT} have proved that pseudo labeling is very helpful to remedy this issue. Like such work, we further propose Neighborhood Expansion to make full use of the unlabeled samples in the target domain. Concretely, by pseudo labeling, we assign a pseudo-label with the highest predicted probability to each unlabeled sample in the target domain at the beginning of each training epoch, as long as its confidence is greater than a certain threshold. A large number of pseudo-labeled samples in the target domain provides more diversified representation learning information to improve the generalization ability of the classifier within the target domain.

Technically, we pre-define a confidence threshold $\tau$, and an unlabeled target domain sample is assigned a pseudo-label $\hat{y}_i^u=\argmax{(p(x_i^u))}$ when the probability (confidence) score of its predicted class label is larger than $\tau$.
Therefore, we select pseudo-labeled target domain samples from $\mathcal{D}_{u}$, merge them into the labeled target domain set $\mathcal{D}_{l}$, and finally form a new target domain set $\mathcal{D}_{l}^{\prime}$. \textcolor{\mycolor}{Here, it is essential to observe that these pseudo-labeled target samples are not excluded from $\mathcal{D}_{u}$. Furthermore,}
$\mathcal{D}_{l}^{\prime}$ is used not only in equations of this section, but also in Equation (\ref{Equation:SDM}) and Equation (\ref{Equation:MDM}) of Section~\ref{Subsection:IDM}.

To make model predictions at unlabeled target domain samples more confident, we introduce {\bf Self-Regularization} and {\bf Pairwise Approaching} to reduce the uncertainty of the predictions, achieving the goal of this by minimizing the following loss,
\begin{equation}
\label{loss:NE}
    {\mathcal{L}}_{\bm{NE}}(\theta;{\mathcal{D}_{l}^{\prime}, \mathcal{D}_{u}}) = {\mathcal{L}}_{\bm{psr}}(\theta;{\mathcal{D}_{u}})
     + {\mathcal{L}}_{\bm{nsr}}(\theta;{\mathcal{D}_{u}}) + {\mathcal{L}}_{\bm{pa}}(\theta;\mathcal{D}_{u}, \mathcal{D}_{l}^{\prime}).
\end{equation}

\subsubsection{Self-Regularization}

Self-Regularization allows the network to learn knowledge from a sample itself. In our proposed method, Self-Regularization includes positive self-regularization learning ({\bf PSR}) and negative self-regularization learning ({\bf NSR}). We employ PSR and NSR to handle unlabeled target domain samples whose predicted probability scores are above and below the confidence threshold $\tau$ respectively. Specifically, different from traditional self-training techniques~\cite{wei2021theoretical, Xie_2020_CVPR, Pan_2020_CVPR}, our {\bf PSR} only operates on unlabeled target domain samples that achieve high predicted probabilities since they are more likely to be assigned with correct pseudo-labels. In addition, similar to ~\cite{NEURIPS2020_06964dce}, the proposed PSR expects an augmented image to have the same model output as its original version. We enforce positive self-regularization learning with a supervised cross-entropy loss as follows,
\begin{equation}
    \begin{split}
        {\mathcal{L}}_{\bm{psr}}(\theta;{\mathcal{D}_{u}})= -\frac{1}{\sum_{x_i^u\in{\mathcal{U}}}\mu_{i}}\sum_{x_i^u\in{\mathcal{U}}}\mu_{i}\cdot{p_y(\hat{y}}_i^{u})\log(p(x_i^u+\delta)),
    \end{split}
\end{equation}
where $\hat{y}_i^u=\argmax{(p(x_i^u))}$ indicates the pseudo-label of a sample $x_i^u$, and $p(x_i^u)$ represents the prediction at an unlabeled image $x_i^u$. Also, $p(x_i^u+\delta)$ represents the prediction at a transformed image $x_i^u+\delta$, where we use RandAugment~\cite{cubuk2020randaugment} to add a perturbation $\delta$ to the original image $x_i^u$. Moreover, $\mu_{i}=\mathds{1}\{{\max(p(x_i^u))\ge\tau}\}$ indicates the probabilistic confidence score of the predicted label of a sample $x_i^u$ should be larger than $\tau$, and $\mathds{1}\{\cdot\}$ is an indicator function. As illustrated in Figure~\ref{Figure-NE}(a), PSR can strengthen the robustness of the model so as to implicitly achieve higher confidence at an unlabeled sample in the target domain.

For unlabeled target domain samples whose probability (confidence) scores associated with their predicted class labels are below the confidence threshold $\tau$, assigning them hard pseudo-labels could easily confuse the model with incorrect label information. As training proceeds, the classifier would gradually fit the assigned noisy pseudo-labels, giving rise to performance degradation. Therefore, as like~\cite{kim2019nlnl}, we here intend to employ {\bf NSR} to assign a hard ``complementary'' label to an unlabeled sample with a low confidence score in the target domain. Concretely, a complementary label is determined by the index of the minimum component of the sample's predicted probability vector. In this case, the complementary label means that the sample has the maximum probability of not belonging to the corresponding class. To this end, we optimize the model via an NSR loss, i.e., ${\mathcal{L}}_{\bm{nsr}}$, to make the probabilities of the predicted complementary class labels of unlabeled target domain samples farther away from $1$ but closer to $0$ as follows:
\begin{equation}
    {\mathcal{L}}_{\bm{nsr}}(\theta;{\mathcal{D}_{u}})=
    -\frac{1}{\sum_{x_i^u\in{\mathcal{U}}}\mu_{i}^{\prime}}\sum_{x_i^u\in{\mathcal{U}}}\mu_{i}^{\prime}\cdot{p_y(\bar{y}}_i^{u})\log(1-p(x_i^u)),
\end{equation}
where $\bar{y}_i^{u}=\argmax{(1-p(x_i^u))}$ and $\mu_{i}^{\prime}=\mathds{1}\{{\max(p(x_i^u))<\tau}\}$ indicates the highest predicted probability score of $x_i^u$ is less than $\tau$. Note that NSR differs from~\cite{kim2019nlnl} since a complementary label in~\cite{kim2019nlnl} is selected from all class labels at random except for the predicted class label corresponding to the highest confidence score of a sample. As shown in Figure~\ref{Figure-NE}(b), after model optimization with the NSR loss, the probability score of the predicted complementary class label of a sample has been almost decreased to 0, but the predicted probabilities of other classes are increased.

\subsubsection{Pairwise Approaching}

Pairwise Approaching ({\bf PA}) is introduced to make feature representations in the target domain more compact by driving unlabeled samples closer to those labeled ones. During the training process, model predictions at more and more unlabeled target domain samples have a confidence level exceeding the threshold $\tau$. 
Given this observation, we draw on the idea of contrastive learning~\cite{NEURIPS2020_492114f6, chen2020simple, zbontar2021barlow} and introduce binary cross-entropy as a loss term to make those unlabeled samples with high confidence approach those labeled target domain samples of the same category to learn more compact features. 
The loss for Pairwise Approaching is formulated as follows:
\begin{equation}
\label{loss_pa}
    \begin{split}
        {\mathcal{L}}_{\bm{pa}}(\theta;{\mathcal{D}_{u}},{\mathcal{D}_{l}^{\prime}})=\quad\quad\quad\quad&\\
        -\frac{1}{\sum_{x_i^u\in{\mathcal{U}}}\mu_{i}}\sum_{x_i^u\in{\mathcal{D}_{u}}}\sum_{x_j^l\in{\mathcal{D}_{l}^{\prime}}}&\mu_{i}\cdot[\nu_{ij}\log(\mathbf{p}_i^{\mathsf{T}}\mathbf{p}_j) \\
        +(1-\nu_{ij})\log(1-&\mathbf{p}_i^{\mathsf{T}}\mathbf{p}_j)],
    \end{split}
\end{equation}
where $\mathbf{p}_{i}=p(x_i^l)$ and $\mathbf{p}_{j}=p(x_j^u)$ represent the predictions at a labeled (or pseudo-labeled) target image $x_i^l$ and an unlabeled target image $x_j^u$, respectively. Also, $\nu_{ij}= \mathds{1}\{{\hat{y}_i^u=y_j^l}\}$ denotes a pairwise label that indicates whether $x_i^u$ and $x_j^l$ have the same class label. As shown in Figure~\ref{Figure-NE}(c), this loss enables the selected unlabeled target domain data to obtain higher predicted probability scores so that the model produces predictions with lower entropy but higher confidence for them. 

\begin{algorithm}[!t]
    \small	
    \DontPrintSemicolon
    \SetNoFillComment
    
    \textbf{Input:} Labeled samples $\mathcal{D}_{s}$, labeled samples $\mathcal{D}_{l}$, unlabeled samples $\mathcal{D}_{u}$, confidence threshold $\tau$, number of training epochs $\mathcal{T}$
    
    \textbf{Output:} Optimal model parameters $\theta$
    
    \For {$epoch=1,2,\ldots,\mathcal{T}$}{
    
        \For {$x\in\mathcal{D}_{u}$}{
        
            $\hat{y}=\argmax{(p(x))}$;
            
            $\mathcal{D}_{l}^{\prime}\leftarrow\mathcal{D}_{l}^{\prime}\cup\{(x,\hat{y})|\max(p(x))\ge\tau\}$;
            
      }
            
            $\mathcal{D}_{l}^{\prime}\leftarrow\mathcal{D}_{l}\cup\mathcal{D}_{l}^{\prime}$;
            
            Randomly sample mini-batches $\mathcal{B}_{s}\subset\mathcal{D}_{s} \mathcal{B}_{l}\subset\mathcal{D}_{l}, \mathcal{B}_{l}^{\prime}\subset\mathcal{D}_{l}^{\prime} and \mathcal{B}_{u}\subset\mathcal{D}_{u}$;

            Update model parameters $\theta$ by applying SGD with the overall loss, ${\mathcal{L}}=\mathcal{L}_{\bm{sup}}(\theta;{\mathcal{B}_{s}},{\mathcal{B}_{l}})+\beta{\mathcal{L}}_{\bm{IDM}}(\theta;{\mathcal{B}_{s}, \mathcal{B}_{l}^{\prime}})+\gamma{\mathcal{L}}_{\bm{NE}}(\theta;{\mathcal{B}_{l}^{\prime}, \mathcal{B}_{u}})$.
}
    \caption{Pseudo-code of IDMNE.}
\label{PseudoCode:IDMNE}
\end{algorithm}

\subsection{Overall Loss Function}

The overall training procedure of our proposed method is described in Algorithm~\ref{PseudoCode:IDMNE}. At the beginning of each epoch, we first use the pseudo labeling scheme to assign pseudo-labels to a subset of unlabeled samples in $\mathcal{D}_{u}$ whose confidence score of the predicted class label is larger than the threshold $\tau$. Then, we update $\mathcal{D}_{l}$ to include the pseudo-labeled target domain samples and form $\mathcal{D}_{l}^{\prime}$. Next, four mini-batches $\mathcal{B}_{s}$, $\mathcal{B}_{l}$, $\mathcal{B}_{l}^{\prime}$ and $\mathcal{B}_{u}$ are assembled by means of random sampling from $\mathcal{D}_{s}$, $\mathcal{D}_{l}$, $\mathcal{D}_{l}^{\prime}$ and $\mathcal{D}_{u}$.
Finally, the overall loss function for model training can be formulated as follows,
\begin{equation}
    \begin{split}
        {\mathcal{L}}=\mathcal{L}_{\bm{sup}}(\theta;{\mathcal{B}_{s}},{\mathcal{B}_{l}})+\beta{\mathcal{L}}_{\bm{IDM}}(\theta;{\mathcal{B}_{s}, \mathcal{B}_{l}^{\prime}})+\gamma{\mathcal{L}}_{\bm{NE}}(\theta;{\mathcal{B}_{l}^{\prime}, \mathcal{B}_{u}}),
    \end{split}
\end{equation}
where $\beta$ and $\gamma$ are two hyper-parameters that trade-offs the losses. In all experiments, we set $\beta=1.0$ and $\gamma=0.1$, respectively. In order to mitigate the influence exerted by under-confident unlabeled samples on the model, we have discovered that configuring a smaller $\gamma$ proves advantageous for the acquisition of a more optimal adaptation model. 

\section{Experiments}

To demonstrate the advantages of the proposed IDMNE method, we have conducted extensive experiments on multiple widely-used benchmarks, namely \texttt{DomainNet}~\cite{PengXingchao2018MMfM}, \texttt{Office}\texttt{-}\texttt{Home}~\cite{VenkateswaraHemanth2017DHNf} and \texttt{Office}\texttt{-}\texttt{31}~\cite{saenko2010adapting}. We first briefly introduce the experimental setups, such as the details of evaluation datasets and the corresponding evaluation protocols. Then, we perform comprehensive comparisons to verify the superiority of our IDMNE over existing state-of-the-art methods. Finally, we have performed detailed ablation studies to demonstrate the contribution of each component within IDMNE. Note that we implemented the proposed method using PyTorch\footnote{https://pytorch.org/}, a popular platform for deep learning, and run all experiments on an NVIDIA GeForce 1080Ti GTX GPU.

\subsection{Datasets}

We evaluate the proposed approach on three widely used SSDA benchmark datasets: \texttt{DomainNet}~\cite{PengXingchao2018MMfM}, \texttt{Office-Home}~\cite{VenkateswaraHemanth2017DHNf} and \texttt{Office-31}~\cite{saenko2010adapting}. Following prior algorithms, the number of labeled target domain samples is set to 1 shot or 3 shots per class.

\texttt{DomainNet} is a standard benchmark dataset for multi-source domain adaptation with a large scale of 0.6 million images. \texttt{DomainNet} has 6 domains and each domain consists of 345 categories. Following~\cite{saito2019semi}, we adopt only 4 domains, namely Real: R, Clipart: C, Painting: P, and Sketch: S, and 126 categories to validate the proposed framework for adaptation.
As in~\cite{saito2019semi}, we also choose seven adaptation scenarios for performance comparisons. 

\texttt{Office-Home} is another popular SSDA benchmark dataset used to evaluate the proposed method with several challenging adaptation scenarios. This dataset involves approximately 65 classes. There are 4 domains, including Real: R, Clipart: C, Art: A, and Product: P. To be fair, we follow previous SSDA work~\cite{saito2019semi, kim2020attract, li2021cross} to carry out 12 adaptation scenarios on this dataset.

\texttt{Office-31} is a small-scale benchmark dataset used for SSDA evaluation. This dataset has three domains, including DSLR: W, Webcam: W, and Amazon: A, and 31 classes.

\subsection{Experimental Protocols}
\label{Subsection:Protocols}

Following~\cite{saito2019semi, YangLuyu2020DCwT, ELP, li2021cross}, we adopt AlexNet~\cite{alexnet} and ResNet-34~\cite{resnet34} as network backbones for evaluation on \texttt{DomainNet} while using AlexNet, VGG-16~\cite{vgg16} and ResNet-34 on \texttt{Office-Home}. We only use AlexNet on \texttt{Office-31}. To achieve a fair comparison, we first initialize the networks with pre-trained weights from ImageNet~\cite{imagenet} and replace the last layer with a prototypical classifier. This prototypical classifier has an unbiased linear layer, which is initialized using random parameters and a temperature $T=0.05$. We perform model training using Stochastic Gradient Descent (SGD) with a momentum of 0.9 and a weight decay of $5\times10^{-4}$. Moreover, the initial learning rate is set to $\eta_{0}=0.001$ and is updated by following the rule in~\cite{saito2019semi} as model training iterates, i.e., $\eta_{t}=\frac{\eta_{0}}{(1+0.0001\times{t})^{0.75}}$, 
where $\eta_{t}$ denotes the learning rate at $t$-th iteration.
 
At the start of each iteration, we randomly sample four mini-batches $\mathcal{B}_{s}\subset\mathcal{D}_{s}$, $\mathcal{B}_{l}\subset\mathcal{D}_{l}$, $\mathcal{B}_{l}^{\prime}\subset\mathcal{D}_{l}^{\prime}$ and $\mathcal{B}_{u}\subset\mathcal{D}_{u}$, where $\mathcal{D}_{l}^{\prime}$ contains labeled samples from $\mathcal{D}_{l}$ as well as pseudo-labeled samples obtained from $\mathcal{D}_{u}$. The mini-batch size of $\mathcal{B}_{s}$, $\mathcal{B}_{l}$, $\mathcal{B}_{l}^{\prime}$ and $\mathcal{B}_{u}$ are set to 24, 24, 24 and 48, respectively when ResNet-34 or VGG-16 is used (32, 32, 32 and 64 for AlexNet). Before being fed into the network, the size of the input images is first resized to $256\times256$, then the images are augmented using the random horizontal flip and random crop ($224\times224$ for VGG-16 and ResNet-34, and $227\times227$ for AlexNet). We finally subtract the per-pixel image mean of the dataset from all images. In addition, the model is trained with 150 epochs (7500 iterations) on $\texttt{DomainNet}$, 100 epochs (2500 iterations) on $\texttt{Office-Home}$, and 100 epochs (1500 iterations) on $\texttt{Office-31}$, i.e., $\mathcal{T}=150$ for \texttt{DomainNet} and $\mathcal{T}=100$ for \texttt{Office-Home} and \texttt{Office-31}. As well, $\tau$ and $\alpha$ are set to 0.95 and 2.0, respectively.
 
\noindent\textcolor{\mycolor}{
\textbf{Setting the Hyper-parameters $\alpha$, $\beta$, $\gamma$, and $\tau$.} We set the value of $\alpha$ to 1.0, which is commonly used in Mixup operations \cite{zhang2018mixup, pmlr-v97-verma19a}. For $\tau$, we decided on a value of 0.95, following the common practice in pseudo-labeling techniques \cite{NEURIPS2020_06964dce}.  To determine suitable values for $\beta$ and $\gamma$, we conducted hyper-parameter selection using the  ``$R \rightarrow  S$'' case on the \texttt{DomainNet} dataset, using ResNet-34 and the 3-shot setup. We then applied these chosen settings to other benchmark datasets and adaptation scenarios, ensuring the reproducibility and simplicity of our proposed algorithm. For the details of choosing of $\beta$ and $\gamma$, similarly to MME \cite{saito2019semi}, we first selected three labeled examples as the validation set for the target domain. We utilized these validation examples to choose those hyper-parameters through a grid search. During this process, we conducted the experiment by fixing the value of $\beta$ to adjust $\gamma$, and then fixing  $\gamma$ to adjust $\beta$. Finally, we chose the values of $\beta=1.0$ and $\gamma=0.1$ based on the highest validation accuracy achieved. 
}

\noindent\textcolor{\mycolor}{
\textbf{Averaged Cluster Centroid Distance.} We propose the averaged Cluster Centroid Distance (ACCD) as a metric, inspired by prior works \cite{ li2021cross}, to evaluate the effectiveness of our Inter-domain Mixup approach in achieving cross-domain feature alignment. ACCD quantifies the distance between the feature clusters of the source and target domains for all classes in the dataset. Specifically, ACCD is calculated as $d_{avg}^{e} = \textbf{average}(\{{d_1^e, d_2^e, \cdots, d_k^e, \cdots, d_K^e}\})$, where $\textbf{average}(\cdot)$ computes the average of the given inputs, and $K$ represents the number of classes in the dataset. Here, $d_k^e$ refers to the pairwise Euclidean distances between the centroids of the feature clusters from the source and target domains for class $k$ at epoch $e$,  each of which should be normalized by the initial distance $d_k^0$, obtained from the initial model with pre-trained weights on ImageNet without any fine-tuning. In general, a smaller ACCD indicates better feature alignment between the source and target domain clusters.
}
 
\begin{table*}[t]
    \centering
    \caption{Comparison results (\%) on 4 domains of \texttt{DomainNet} under the 3-shot setting using AlexNet.  \textcolor{\mycolor}{(Mean accuracy and 95\% confidence interval over five trails)}}
    \tiny
    \begin{tabular}
        {l|>{\centering\arraybackslash}m{0.705cm} >{\centering\arraybackslash}m{0.705cm}>{\centering\arraybackslash}m{0.705cm}>{\centering\arraybackslash}m{0.705cm}>{\centering\arraybackslash}m{0.705cm}>{\centering\arraybackslash}m{0.705cm}>{\centering\arraybackslash}m{0.705cm}|>{\centering\arraybackslash}m{0.705cm}}
        \toprule
        \multicolumn{1}{l|}{Method} & R$\rightarrow$C   & R$\rightarrow$P   & P$\rightarrow$C   & C$\rightarrow$S   & S$\rightarrow$P   & R$\rightarrow$S   & P$\rightarrow$R   & Mean \\
        \midrule
        S+T~\cite{saito2019semi}          & 47.1  & 45.0  & 44.9  & 36.4  & 38.4  & 33.3  & 58.7  & 43.4  \\
        DANN~\cite{saito2019semi}         & 46.1  & 43.8  & 41.0  & 36.5  & 38.9  & 33.4  & 57.3  & 42.4  \\
        MME~\cite{saito2019semi}          & 55.6  & 49.0  & 51.7  & 39.4  & 43.0  & 37.9  & 60.7  & 48.2  \\
        Meta-MME~\cite{li2020online}      & 56.4  & 50.2  & 51.9  & 39.6  & 43.7  & 38.7  & 60.7  & 48.8  \\
        BiAT~\cite{2020Bidirectional}     & 58.6  & 50.6  & 52.0  & 41.9  & 42.1  & 42.0  & 58.8  & 49.4  \\
        APE~\cite{kim2020attract}         & 54.6  & 50.5  & 52.1  & 42.6  & 42.2  & 38.7  & 61.4  & 48.9  \\
        PAC~\cite{mishra2021surprisingly} & 61.7  & 56.9  & 59.8  & 52.9  & 43.9  & 48.2  & 59.7  & 54.7  \\
        Relaxed-cGAN~\cite{luo2021relaxed} & 56.8  & 51.8  & 52.0  & 44.1  & 44.2  & 42.8  & 61.1  & 50.5  \\
        CDAC~\cite{li2021cross}           & 61.4  & 57.5  & 58.9  & 50.7  & 51.7  & 46.7  & 66.8  & 56.2  \\
        IDMNE (Ours)    & \textcolor{\mycolor}{\textbf{63.17±0.22}} & \textcolor{\mycolor}{\textbf{58.96±0.30}} & \textcolor{\mycolor}{\textbf{61.48±0.21}} & \textcolor{\mycolor}{\textbf{54.88±0.77}} & \textcolor{\mycolor}{\textbf{53.58±0.42}} & \textcolor{\mycolor}{\textbf{48.52±0.44}} & \textcolor{\mycolor}{\textbf{67.49±0.22}} & \textcolor{\mycolor}{\textbf{58.30}} \\
        \bottomrule
    \end{tabular}
    \label{Table:Domainnet-Alex-3shot}
\end{table*}

\begin{table*}[t]
    \centering
    \caption{Comparison results (\%) on 4 domains of \texttt{DomainNet} under the 1-shot and 3-shot settings using ResNet-34.  \textcolor{\mycolor}{(Mean accuracy and 95\% confidence interval over five trails)}}
    \resizebox{\linewidth}{!}{
    \begin{tabular}{l|cccccccccccccc|cc}
    \toprule
    \multirow{2}[2]{*}{Method} & \multicolumn{2}{c}{R→C} & \multicolumn{2}{c}{R→P} & \multicolumn{2}{c}{P→C} & \multicolumn{2}{c}{C→S} & \multicolumn{2}{c}{S→P} & \multicolumn{2}{c}{R→S} & \multicolumn{2}{c|}{P→R} & \multicolumn{2}{c}{Mean} \\
          & 1-shot & 3-shot & 1-shot & 3-shot & 1-shot & 3-shot & 1-shot & 3-shot & 1-shot & 3-shot & 1-shot & 3-shot & 1-shot & 3-shot & 1-shot & 3-shot \\
    \midrule
    S+T~\cite{saito2019semi}   & 55.6  & 60.0  & 60.6  & 62.2  & 56.8  & 59.4  & 50.8  & 55.0  & 56.0  & 59.5  & 46.3  & 50.1  & 71.8  & 73.9  & 56.9  & 60.0  \\
    DANN~\cite{saito2019semi}  & 58.2  & 59.8  & 61.4  & 62.8  & 56.3  & 59.6  & 52.8  & 55.4  & 57.4  & 59.9  & 52.2  & 54.9  & 70.3  & 72.2  & 58.4  & 60.7  \\
    MME~\cite{saito2019semi}   & 70.0  & 72.2  & 67.7  & 69.7  & 69.0  & 71.7  & 56.3  & 61.8  & 64.8  & 66.8  & 61.0  & 61.9  & 76.1  & 78.5  & 66.4  & 68.9  \\
    UODA~\cite{qin2021contradictory}  & 72.7  & 75.4  & 70.3  & 71.5  & 69.8  & 73.2  & 60.5  & 64.1  & 66.4  & 69.4  & 62.7  & 64.2  & 77.3  & 80.8  & 68.5  & 71.2  \\
    BiAT~\cite{2020Bidirectional}  & 73.0  & 74.9  & 68.0  & 68.8  & 71.6  & 74.6  & 57.9  & 61.5  & 63.9  & 67.5  & 58.5  & 62.1  & 77.0  & 78.6  & 67.1  & 69.7  \\
    APE~\cite{kim2020attract}   & 70.4  & 76.6  & 70.8  & 72.1  & 72.9  & 76.7  & 56.7  & 63.1  & 64.5  & 66.1  & 63.0  & 67.8  & 76.6  & 79.4  & 67.6  & 71.7  \\
    PAC~\cite{mishra2021surprisingly}   & 74.9  & 78.6  & 73.0  & 74.3  & 72.6  & 76.0  & 65.8  & 69.6  & 67.9  & 69.4  & 68.7  & 70.2  & 76.7  & 79.3  & 71.4  & 73.9  \\
    ELP~\cite{ELP}   & 72.8  & {74.9} & 70.8  & 72.1  & 72.0  & 74.4  & 59.6  & 64.3  & 66.7  & 69.7  & 63.3  & 64.9  & 77.8  & 81.0  & 69.0  & 71.6  \\
    DECOTA~\cite{YangLuyu2020DCwT} & 79.1  & 80.4  & 74.9  & 75.2  & 76.9  & 78.7  & 65.1  & 68.6  & 72.0  & 72.7  & 69.7  & 71.9  & 79.6  & 81.5  & 73.9  & 75.6  \\
    CDAC~\cite{li2021cross}  & 77.4  & 79.6  & 74.2  & 75.1  & 75.5  & 79.3  & 67.6  & 69.9  & 71.0  & 73.4  & 69.2  & 72.5  & 80.4  & 81.9  & 73.6  & 76.0  \\
    UODAv2~\cite{qin2022semi} & 77.0  & 79.4  & 75.4  & 76.7  & 75.5  & 78.3  & 66.5  & 70.2  & 72.1  & 74.2  & 70.9  & 72.1  & 79.7  & 82.3  & 73.9  & 76.2  \\
    IDMNE (Ours) & \textcolor{\mycolor}{\textbf{79.56±0.21}} & \textcolor{\mycolor}{\textbf{80.81±0.21}} & \textcolor{\mycolor}{\textbf{75.95±0.30}} & \textcolor{\mycolor}{\textbf{76.88±0.11}} & \textcolor{\mycolor}{\textbf{79.43±0.17}} & \textcolor{\mycolor}{\textbf{80.29±0.13}} & \textcolor{\mycolor}{\textbf{71.69±0.73}} & \textcolor{\mycolor}{\textbf{72.22±0.35}} & \textcolor{\mycolor}{\textbf{75.35±0.48}} & \textcolor{\mycolor}{\textbf{75.39±0.12}} & \textcolor{\mycolor}{\textbf{73.48±0.64}} & \textcolor{\mycolor}{\textbf{73.92±0.28}} & \textcolor{\mycolor}{\textbf{82.14±0.67}} & \textcolor{\mycolor}{\textbf{82.80±0.06}} & \textcolor{\mycolor}{\textbf{76.80}} & \textcolor{\mycolor}{\textbf{77.47}}\\
    \bottomrule
    \end{tabular}
}
  \label{TableL:DomainNet-Res}
\end{table*}

\begin{table*}[t]
  \centering
  \caption{Comparison results (\%) on 4 domains of \texttt{Office-Home} under the 3-shot setting using AlexNet.  \textcolor{\mycolor}{(Mean accuracy and 95\% confidence interval over five trails)}}
  \resizebox{\linewidth}{!}{
    \begin{tabular}{l|cccccccccccc|c}
    \toprule
    Method & R→C   & R→P   & R→A   & P→R   & P→C   & P→A   & A→P   & A→C   & A→R   & C→R   & C→A   & C→P   & Mean \\
    \midrule
    S+T~\cite{saito2019semi}   & 44.6  & 66.7  & 47.7  & 57.8  & 44.4  & 36.1  & 57.6  & 38.8  & 57.0  & 54.3  & 37.5  & 57.9  & 50.0 \\
    DANN~\cite{saito2019semi}  & 47.2  & 66.7  & 46.6  & 58.1  & 44.4  & 36.1  & 57.2  & 39.8  & 56.6  & 54.3  & 38.6  & 57.9  & 50.3 \\
    ENT~\cite{saito2019semi}   & 44.9  & 70.4  & 47.1  & 60.3  & 41.2  & 34.6  & 60.7  & 37.8  & 60.5  & 58.0  & 31.8  & 63.4  & 50.9 \\
    MME~\cite{saito2019semi}   & 51.2  & 73.0  & 50.3  & 61.6  & 47.2  & 40.7  & 63.9  & 43.8  & 61.4  & 59.9  & 44.7  & 64.7  & 55.2 \\
    Meta-MME~\cite{li2020online} & 50.3  & -     & -     & -     & 48.3  & 40.3  & -     & 44.5  & -     & -     & 44.5  & -     & - \\
    BiAT~\cite{2020Bidirectional}  & -     & -     & -     & -     & -     & -     & -     & -     & -     & -     & -     & -     & 56.4 \\
    APE~\cite{kim2020attract}   & 51.9  & 74.6  & 51.2  & 61.6  & 47.9  & 42.1  & 65.5  & 44.5  & 60.9  & 58.1  & 44.3  & 64.8  & 55.6 \\
    PAC~\cite{mishra2021surprisingly}   & \textbf{58.9} & 72.4  & 47.5  & 61.9  & \textbf{53.2} & 39.6  & 63.8  & \textbf{49.9} & 60.0  & 54.5  & 36.3  & 64.8  & 55.2 \\
    CDAC~\cite{li2021cross}  & 54.9  & 75.8  & 51.8  & 64.3  & 51.3  & \textbf{43.6} & 65.1  & 47.5  & \textbf{63.1} & 63.0  & \textbf{44.9} & 65.6  & 56.8 \\
    IDMNE (Ours) &  \textcolor{\mycolor}{{55.91±0.38}} & \textcolor{\mycolor}{\textbf{76.91±0.56}} & \textcolor{\mycolor}{\textbf{51.98±0.25}} & \textcolor{\mycolor}{\textbf{65.79±0.22}} & \textcolor{\mycolor}{{53.00±0.72}} & \textcolor{\mycolor}{{43.48±0.69}} & \textcolor{\mycolor}{\textbf{66.28±0.31}} & \textcolor{\mycolor}{{47.20±0.78}} & \textcolor{\mycolor}{{61.10±0.42}} & \textcolor{\mycolor}{\textbf{63.69±0.58}} & \textcolor{\mycolor}{{42.83±0.16}} & \textcolor{\mycolor}{\textbf{65.97±0.42}} & \textcolor{\mycolor}{\textbf{57.84}} \\
    \bottomrule
    \end{tabular}
}
  \label{TableL:Office-Alex-3shot}
\end{table*}

\begin{table*}[t]
  \centering
  \caption{Comparison results (\%) on 4 domains of \texttt{Office-Home} under the 1-shot and 3-shot settings using VGG-16.  \textcolor{\mycolor}{(Mean accuracy and 95\% confidence interval over five trails)}}
  \resizebox{\linewidth}{!}{
    \begin{tabular}{l|cccccccccccc|c}
    \toprule
    Method & R→C   & R→P   & R→A   & P→R   & P→C   & P→A   & A→P   & A→C   & A→R   & C→R   & C→A   & C→P   & Mean \\
    \midrule
    \multicolumn{14}{c}{1-shot} \\
    \midrule
    S+T~\cite{saito2019semi}   & 39.5  & 75.3  & 61.2  & 71.6  & 37.0  & 52.0  & 63.6  & 37.5  & 69.5  & 64.5  & 51.4  & 65.9  & 57.4 \\
    DANN~\cite{saito2019semi}  & 52.0  & 75.7  & 62.7  & 72.7  & 45.9  & 51.3  & 64.3  & 44.4  & 68.9  & 64.2  & 52.3  & 65.3  & 60.0 \\
    ENT~\cite{saito2019semi}   & 23.7  & 77.5  & 64.0  & 74.6  & 21.3  & 44.6  & 66.0  & 22.4  & 70.6  & 62.1  & 25.1  & 67.7  & 51.6 \\
    MME~\cite{saito2019semi}   & 49.1  & 78.7  & 65.1  & 74.4  & 46.2  & 56.0  & 68.6  & 45.8  & 72.2  & 68.0  & 57.5  & 71.3  & 62.7 \\
    UODA~\cite{qin2021contradictory}  & 49.6  & 79.8  & 66.1  & 75.4  & 45.5  & 58.8  & 72.5  & 43.3  & \textbf{73.3}  & 70.5  & 59.3  & 72.1  & 63.9 \\
    ELP~\cite{ELP}   & 49.2  & 79.7  & 65.5  & 75.3  & 46.7  & 56.3  & 69.0  & 46.1  & 72.4  & 68.2  & \textbf{67.4} & 71.6  & 63.1 \\
    DECOTA~\cite{YangLuyu2020DCwT} & 47.2  & 80.3  & 64.6  & 75.5  & 47.2  & 56.6  & 71.1  & 42.5  & 73.1  & 71.0  & 57.8  & 72.9  & 63.3 \\
    UODAv2~\cite{qin2022semi} & 51.6  & 80.9  & 66.9  & 75.9  & 49.7  & \textbf{60.5} & 71.0  & 44.9  & {73.2} & 70.6  & 58.7  & 72.8  & 64.7 \\
    IDMNE (Ours) &         \textcolor{\mycolor}{\textbf{52.61±0.92}} & \textcolor{\mycolor}{\textbf{81.75±0.98}} & \textcolor{\mycolor}{\textbf{67.51±0.23}} & \textcolor{\mycolor}{\textbf{77.27±0.06}} & \textcolor{\mycolor}{\textbf{50.67±0.45}} & \textcolor{\mycolor}{{59.70±0.74}} & \textcolor{\mycolor}{\textbf{73.71±0.77}} & \textcolor{\mycolor}{\textbf{49.62±0.06}} & \textcolor{\mycolor}{{72.64±0.41}} & \textcolor{\mycolor}{\textbf{71.42±0.18}} & \textcolor{\mycolor}{{62.52±0.53}} & \textcolor{\mycolor}{\textbf{76.17±1.70}} & \textcolor{\mycolor}{\textbf{66.30}} \\
    \midrule
    \multicolumn{14}{c}{3-shot} \\
    \midrule
    S+T~\cite{saito2019semi}   & 49.6  & 78.6  & 63.6  & 72.7  & 47.2  & 55.9  & 69.4  & 47.5  & 73.4  & 69.7  & 56.2  & 70.4  & 62.9 \\
    DANN~\cite{saito2019semi}  & 56.1  & 77.9  & 63.7  & 73.6  & 52.4  & 56.3  & 69.5  & 50.0  & 72.3  & 68.7  & 56.4  & 69.8  & 63.9 \\
    ENT~\cite{saito2019semi}   & 48.3  & 81.6  & 65.5  & 76.6  & 46.8  & 56.9  & 73.0  & 44.8  & 75.3  & 72.9  & 59.1  & 77.0  & 64.8 \\
    MME~\cite{saito2019semi}   & 56.9  & 82.9  & 65.7  & 76.7  & 53.6  & 59.2  & 75.7  & 54.9  & 75.3  & 72.9  & 61.1  & 76.3  & 67.6 \\
    UODA~\cite{qin2021contradictory}  & 57.6  & 83.6  & 67.5  & 77.7  & 54.9  & 61.0  & 77.7  & 55.4  & \textbf{76.7} & 73.8  & 61.9  & 78.4  & 68.9 \\
    APE~\cite{kim2020attract}   & 56.0  & 81.0  & 65.2  & 73.7  & 51.4  & 59.3  & 75.0  & 54.4  & 73.7  & 71.4  & 61.7  & 75.1  & 66.5 \\
    ELP~\cite{ELP}   & 57.1  & 83.2  & 67.0  & 76.3  & 53.9  & 59.3  & 75.9  & 55.1  & 76.3  & 73.3  & 61.9  & 76.1  & 68.0 \\
    DECOTA~\cite{YangLuyu2020DCwT} & 59.9  & 83.9  & 67.7  & 77.3  & 57.7  & 60.7  & 78.0  & 54.9  & 76.0  & 74.3  & 63.2  & 78.4  & 69.3 \\
    UODAv2~\cite{qin2022semi} & 59.3  & 83.6  & 68.0  & \textbf{78.3} & 56.8  & 61.8  & \textbf{78.6} & 55.7  & 75.3  & 74.0  & 63.3  & 78.9  & 69.5 \\
    IDMNE (Ours) & \textcolor{\mycolor}{\textbf{60.21±0.29}} & \textcolor{\mycolor}{\textbf{84.42±0.59}} & \textcolor{\mycolor}{\textbf{69.33±0.38}} & \textcolor{\mycolor}{{77.92±0.44}} & \textcolor{\mycolor}{\textbf{59.15±1.06}} & \textcolor{\mycolor}{\textbf{62.63±0.81}} & \textcolor{\mycolor}{{77.68±1.16}} & \textcolor{\mycolor}{\textbf{58.24±0.15}} & \textcolor{\mycolor}{{76.68±0.20}} & \textcolor{\mycolor}{\textbf{74.89±0.49}} & \textcolor{\mycolor}{\textbf{64.56±0.41}} & \textcolor{\mycolor}{\textbf{79.27±0.32}} & \textcolor{\mycolor}{\textbf{70.41}} \\
    \bottomrule
    \end{tabular}
}
  \label{TableL:Office-VGG}
\end{table*}

\subsection{Comparison with the State of the Arts}

We perform experimental comparisons between the proposed method and existing state-of-the-art SSDA algorithms, including {\bf S+T}~\cite{saito2019semi}, {\bf DANN}~\cite{ganin2015unsupervised}, {\bf MME}~\cite{saito2019semi}, {\bf UODA}~\cite{qin2021contradictory}, {\bf Meta-MME}~\cite{li2020online}, {\bf BiAT}~\cite{2020Bidirectional}, {\bf APE}~\cite{kim2020attract}, {\bf PAC}~\cite{mishra2021surprisingly}, {\bf ELP}~\cite{ELP}, {\bf DECOTA}~\cite{YangLuyu2020DCwT}, {\bf Relaxed-cGAN}~\cite{luo2021relaxed}, {\bf CDAC}~\cite{li2021cross}, and {\bf UODAv2}~\cite{qin2022semi}. The results of {\bf S+T} and {\bf DANN} are borrowed from \cite{saito2019semi}. {\bf S+T} trains the model using a cross-entropy loss over the labeled source and target domain data only. {\bf DANN} is modified from~\cite{ganin2015unsupervised}, and the model is trained with labeled source domain data, unlabeled target domain data, and a small amount of labeled target domain data. In addition, {\bf MME}, {\bf Meta-MME}, {\bf UODA} and {\bf CDAC} are adversarial learning based approaches while {\bf APE} and {\bf Relaxed-cGAN} mainly focus on minimizing cross-domain discrepancy measures and rely on image style transfer, while {\bf UODAv2} is the extension of {\bf UODA}. Furthermore, to address the SSDA problem, {\bf Meta-MME}, {\bf BiAT}, {\bf PAC} and {\bf DECOTA} adapt previous techniques, such as meta-learning~\cite{schmidhuber1992learning, schmidhuber1997shifting}, VAT~\cite{miyato2018virtual}, FixMatch~\cite{NEURIPS2020_06964dce}, Co-training~\cite{blum1998combining, balcan2004co, chen2011automatic}, and so on.
\textcolor{\mycolor}{
Tables~\ref{Table:Domainnet-Alex-3shot}-\ref{TableL:Office31-Alex-3shot} have listed the mean accuracy and 95\% confidence interval over five trials in each experiment of different adaptation scenarios. The results demonstrate that across different datasets using various network backbones under either 1-shot or 3-shot setups, the average performance for all adaptation scenarios achieves higher than all existing state-of-the-art algorithms, illustrating the superiority of the proposed approach in handling the SSDA task.
Additionally, it can be observed that, in the majority of individual adaptation cases,  the results show that our average accuracy with the lower bound still achieves the best. This shows that the proposed method can obtain statistically significant performance improvement over most of the existing best-performing methods. 
}

\begin{table*}[t]
  \centering
 \caption{Comparison results (\%) on 4 domains of \texttt{Office-Home} under the 3-shot setting using ResNet-34.  \textcolor{\mycolor}{(Mean accuracy and 95\% confidence interval over five trails)}}
    \tiny
    \resizebox{\linewidth}{!}{
    \begin{tabular}{l|cccccccccccc|c}
    \toprule
    Method & R→C   & R→P   & R→A   & P→R   & P→C   & P→A   & A→P   & A→C   & A→R   & C→R   & C→A   & C→P   & Mean \\
    \midrule
    S+T~\cite{saito2019semi}   & 55.7  & 80.8  & 67.8  & 73.1  & 53.8  & 63.5  & 73.1  & 54.0  & 74.2  & 68.3  & 57.6  & 72.3  & 66.2 \\
    DANN~\cite{saito2019semi}  & 57.3  & 75.5  & 65.2  & 69.2  & 51.8  & 56.6  & 68.3  & 54.7  & 73.8  & 67.1  & 55.1  & 67.5  & 63.5 \\
    ENT~\cite{saito2019semi}   & 62.6  & 85.7  & 70.2  & 79.9  & 60.5  & 63.9  & 79.5  & 61.3  & 79.1  & 76.4  & 64.7  & 79.1  & 71.9 \\
    MME~\cite{saito2019semi}   & 64.6  & 85.5  & 71.3  & 80.1  & 64.6  & 65.5  & 79.0  & 63.6  & 79.7  & 76.6  & 67.2  & 79.3  & 73.1 \\
    Meta-MME~\cite{li2020online} & 65.2  & -     & -     & -     & 64.5  & 66.7  & -     & 63.3  & -     & -     & 67.5  & -     & - \\
    APE~\cite{kim2020attract}   & 66.4  & 86.2  & 73.4  & 82.0  & 65.2  & 66.1  & 81.1  & 63.9  & 80.2  & 76.8  & 66.6  & 79.9  & 74.0 \\
    Relaxed-cGAN~\cite{luo2021relaxed} & 68.4  & 85.5  & 73.8  & 81.2  & \textbf{68.1} & 67.9  & 80.1  & 64.3  & 80.1  & 77.5  & 66.3  & 78.3  & 74.2 \\
    DECOTA~\cite{YangLuyu2020DCwT} & 70.4  & 87.7  & 74.0  & 82.1  & 68.0  & \textbf{69.9} & 81.8  & 64.0  & 80.5  & 79.0  & 68.0  & \textbf{83.2} & 75.7 \\
    CDAC~\cite{li2021cross}  & 67.8  & 85.6  & 72.2  & 81.9  & 67.0  & 67.5  & 80.3  & 65.9  & \textbf{80.6} & \textbf{80.2} & 67.4  & 81.4  & 74.2 \\
    IDMNE (Ours) &  \textcolor{\mycolor}{\textbf{71.73±0.56}} &
                    \textcolor{\mycolor}{\textbf{88.09±0.33}} &
                    \textcolor{\mycolor}{\textbf{75.16±0.17}} &
                    \textcolor{\mycolor}{\textbf{82.68±0.26}} &
                    \textcolor{\mycolor}{{67.58±0.23}} &
                    \textcolor{\mycolor}{{68.98±0.28}} &
                    \textcolor{\mycolor}{\textbf{82.41±0.32}} &
                    \textcolor{\mycolor}{\textbf{66.39±0.68}} &
                    \textcolor{\mycolor}{{79.32±0.18}} &
                    \textcolor{\mycolor}{{79.49±0.52}} &
                    \textcolor{\mycolor}{\textbf{69.10±0.79}} &
                    \textcolor{\mycolor}{{83.08±0.77}} &
                    \textcolor{\mycolor}{\textbf{76.17}} \\
    \bottomrule
    \end{tabular}
}
   \label{TableL:Office-RES-3shot}
\end{table*}

\begin{table}[t]
  \centering
  \caption{Comparison results (\%) on \texttt{Office-31} under the 3-shot setting  using AlexNet. \textcolor{\mycolor}{(Mean accuracy and 95\% confidence interval over five trails)}}
  \tiny
  \begin{tabular}{l|cc|c}
    \toprule
    Method & W$\rightarrow$A   & D$\rightarrow$A   & Mean \\
    \midrule
    S+T~\cite{saito2019semi}   & 61.2  & 62.4  & 61.8  \\
    DANN~\cite{saito2019semi}  & 64.4  & 65.2  & 64.8  \\
    ENT~\cite{saito2019semi}   & 64.0  & 66.2  & 65.1  \\
    MME~\cite{saito2019semi}   & 67.3  & 67.8  & 67.6  \\
    BiAT~\cite{2020Bidirectional}  & 68.2  & 68.5  & 68.4  \\
    APE~\cite{kim2020attract} & 67.6  & 69.0  & 68.3  \\
    CDAC~\cite{li2021cross}  & 70.1  & 70.0  & 70.0  \\
    IDMNE (Ours) & \textcolor{\mycolor}{\textbf{71.03±0.34}} & \textcolor{\mycolor}{\textbf{71.32 ±0.47}} & \textcolor{\mycolor}{\textbf{71.18}} \\
    \bottomrule
    \end{tabular}
  \label{TableL:Office31-Alex-3shot}
\end{table}

\noindent
\textbf{Results on DomainNet.} Comparison results between our method and previous SSDA algorithms on \texttt{DomainNet} are shown in Tables~\ref{Table:Domainnet-Alex-3shot}-\ref{TableL:DomainNet-Res}. On this dataset, we carry out experiments under the 3-shot setting using AlexNet and ResNet-34, and under the 1-shot setting using ResNet-34. The average performance of the proposed method exceeds that of previous algorithms by large margins under all settings. This demonstrates that our method can perform well in diverse adaptation scenarios defined on \texttt{DomainNet}. Specifically, as shown in Table~\ref{Table:Domainnet-Alex-3shot}, our algorithm improves the mean accuracy achieved by the existing best-performing algorithm, i.e., {\bf CDAC}, by \textcolor{\mycolor}{2.10}\%, under the 3-shot setting using AlexNet. Moreover, Table~\ref{TableL:DomainNet-Res} also shows that the proposed algorithm achieves the highest mean accuracy and exceeds {\bf UODAv2} by \textcolor{\mycolor}{2.90}\% and \textcolor{\mycolor}{1.27}\%, respectively,  under the 1-shot and 3-shot settings using ResNet-34.

\noindent
\textbf{Results on Office-Home.} To further validate the effectiveness of IDMNE, we compare our method with existing algorithms on the smaller \texttt{Office-Home} benchmark. For fair comparisons, we conduct experiments only under the 3-shot setting when AlexNet or ResNet-34 is the backbone network but report the experimental results under both 1-shot and 3-shot settings when VGG-16 is used as the backbone. Comparison results on \texttt{Office-Home} are shown in Tables~\ref{TableL:Office-Alex-3shot}-\ref{TableL:Office-RES-3shot}, indicating that the proposed method again achieves the best average performance in all adaptation cases under all settings. In particular, Table~\ref{TableL:Office-VGG} shows that in comparison to the highest accuracy achieved by previous algorithms, the average performance gain of our proposed method is respectively \textcolor{\mycolor}{1.60}\% and \textcolor{\mycolor}{0.91}\% under the 1-shot and 3-shot settings when VGG-16 is the backbone. Furthermore, our method also performs well on \texttt{Office-Home} with \textcolor{\mycolor}{1.04}\% and \textcolor{\mycolor}{0.47}\% performance boosts  when AlexNet and ResNet-34 work as the backbone networks, respectively.

\noindent
\textbf{Results on Office-31.} We also evaluate IDMNE using the adaptation tasks defined on \texttt{Office-31}. For a fair comparison with previous algorithms, only AlexNet is adopted as the backbone in the experiments on \texttt{Office-31}. Table~\ref{TableL:Office31-Alex-3shot} shows the comparison result under the 3-shot setting. The proposed method achieves a mean accuracy of 70.9\%, which boosts the previous best-performing algorithm, {\bf CDAC}, by 0.9\%, suggesting that our algorithm also works well on small and relatively simple datasets.

\begin{table*}[t]
    \centering
    \caption{Ablation study of Inter-domain Mixup on \texttt{DomainNet} under the 3-shot setting using ResNet-34.}
    \tiny
    \begin{tabular}{l|>{\centering\arraybackslash}m{0.502cm} >{\centering\arraybackslash}m{0.502cm}>{\centering\arraybackslash}m{0.502cm}>{\centering\arraybackslash}m{0.502cm}>{\centering\arraybackslash}m{0.502cm}>{\centering\arraybackslash}m{0.502cm}c|c}
    \toprule
    \multicolumn{1}{l|}{Method} & \multicolumn{1}{c}{R$\rightarrow$C} & \multicolumn{1}{c}{R$\rightarrow$P} & \multicolumn{1}{c}{P$\rightarrow$C} & \multicolumn{1}{c}{C$\rightarrow$S} & \multicolumn{1}{c}{S$\rightarrow$P} & \multicolumn{1}{c}{R$\rightarrow$S} & \multicolumn{1}{c|}{P$\rightarrow$R} & \multicolumn{1}{c}{Mean} \\
    \midrule
    Baseline1 & 60.0  & 62.2  & 59.4  & 55.0  & 59.5  & 50.1  & 73.9  & 60.0  \\
    Baseline1+SDM & 74.7  & 72.9  & 75.5  & 69.1  & 70.8  & 68.9  & 79.8  & 73.1  \\
    Baseline1+MDM & 76.9  & 73.1  & 76.8  & 69.3  & 71.3  & 69.6  & 80.9  & 74.0  \\
    Baseline1+SDM+MDM & 77.7  & 75.3  & 78.3  & 70.7  & 72.4  & 71.7  & 81.0  & 75.3  \\
    \bottomrule
    \end{tabular}
    \label{TableL:Ablation-IDM}
\end{table*}
 
\begin{table*}[t]
    \centering
    \caption{Ablation study of Neighborhood Expansion on \texttt{DomainNet} under the 3-shot setting using ResNet-34.}
    \tiny
    \begin{tabular}{l|>{\centering\arraybackslash}m{0.502cm} >{\centering\arraybackslash}m{0.502cm}>{\centering\arraybackslash}m{0.502cm}>{\centering\arraybackslash}m{0.502cm}>{\centering\arraybackslash}m{0.502cm}>{\centering\arraybackslash}m{0.502cm}>{\centering\arraybackslash}m{0.502cm}|>{\centering\arraybackslash}m{0.502cm}}
        \toprule
        \multicolumn{1}{l|}{Method} & \multicolumn{1}{c}{R$\rightarrow$C} & \multicolumn{1}{c}{R$\rightarrow$P} & \multicolumn{1}{c}{P$\rightarrow$C} & \multicolumn{1}{c}{C$\rightarrow$S} & \multicolumn{1}{c}{S$\rightarrow$P} & \multicolumn{1}{c}{R$\rightarrow$S} & \multicolumn{1}{c|}{P$\rightarrow$R} & \multicolumn{1}{c}{Mean} \\
        \midrule
        Baseline2 & 77.7  & 75.3  & 78.3  & 70.7  & 72.4  & 71.7  & 81.0  & 75.3  \\
        Baseline2+PSR & 78.5  & 76.0  & 79.2  & 72.0  & 73.1  & 72.2  & 82.2  & 76.2  \\
        Baseline2+PSR+NSR & 79.8  & 76.7  & 79.8  & 71.6  & 74.4  & 73.3  & 82.9  & 76.9  \\
        Baseline2+PSR+NSR+PA (i.e., IDMNE) & 80.7  & 77.0  & 80.6  & 72.1  & 75.2  & 74.2  & 82.7  & 77.5  \\
        \bottomrule
    \end{tabular}
    \label{TableL:Ablation-NE}
\end{table*}

\subsection{Analysis}
\label{Subsection:Analysis}
In this section, we first evaluate the effect of our proposed IDMNE in two aspects, including Inter-domain Mixup and Neighborhood Expansion. Then, further analysis is performed to validate the impact of several important factors w.r.t our approach.

\noindent
\textbf{Ablation Study on Inter-domain Mixup.}
Inter-domain Mixup involves two loss functions, i.e., $\mathcal{L}_{\bm{sdm}}$ and $\mathcal{L}_{\bm{mdm}}$, to achieve cross-domain feature alignment. We conduct variants of Inter-domain Mixup to verify the efficacy of each loss function. We first design a baseline experiment, namely ``Baseline1'', where the model is trained only with $\mathcal{L}_{\bm{sup}}$. Then, adding $\mathcal{L}_{\bm{sdm}}$ and $\mathcal{L}_{\bm{mdm}}$ in turn to ``Baseline1'' forms ``Baseline1+SDM'', ``Baseline1+MDM'' and ``Baseline1+SDM+MDM''. As shown in Table~\ref{TableL:Ablation-IDM}, ``Baseline1+SDM+MDM'' improves ``Baseline1'' by 15.3\% on average while the performance gain of ``Baseline1+SDM'' and ``Baseline1+MDM'' over ``Baseline1'' reaches 13.1\% and 14.0\%, respectively. This means that both $\mathcal{L}_{\bm{sdm}}$ and $\mathcal{L}_{\bm{mdm}}$ improve the performance of ``Baseline1'', and incorporating both into ``Baseline1'' achieves a greater performance improvement, demonstrating the complementarity of the two.
 
\noindent
\textbf{Ablation Study on Neighborhood Expansion.}
Neighborhood Expansion has three loss functions, including $\mathcal{L}_{\bm{psr}}$, $\mathcal{L}_{\bm{nsr}}$ and $\mathcal{L}_{\bm{pa}}$, which are used to further optimize the model trained with $\mathcal{L}_{\bm{sup}}+\mathcal{L}_{\bm{sdm}}+\mathcal{L}_{\bm{mdm}}$ (Called ``Baseline2''). We compare diverse loss functions in Neighborhood Expansion in order to verify the validity of each loss term. For simplification, we design three variants, namely ``Baseline2+PSR'', ``Baseline2+PSR+NSR'' and ``Baseline2+PSR+NSR+PA'', to gradually add $\mathcal{L}_{\bm{psr}}$, $\mathcal{L}_{\bm{nsr}}$ and $\mathcal{L}_{\bm{pa}}$ to the loss function for ``Baseline2''. As shown in Table~\ref{TableL:Ablation-NE}, in comparison to the performance of ``Baseline2'', the gain of each variant in mean accuracy reaches 0.9\%, 1.6\% and 2.2\%, respectively, suggesting that both Self-Regularization and Pairwise Approaching are of significance for the model's performance.
 
\begin{figure*}[t]
    \centering
    \includegraphics[width=0.95\linewidth]{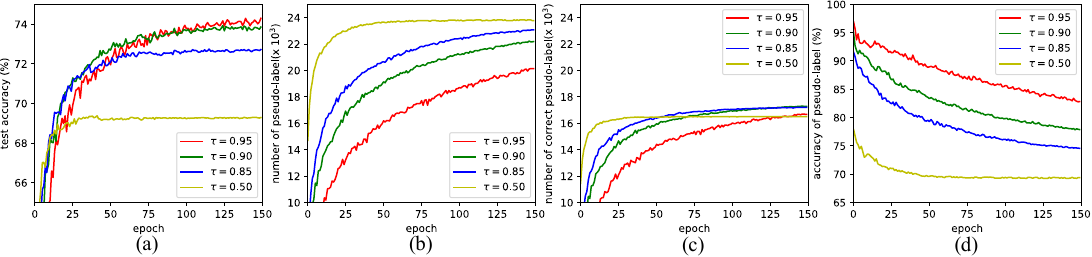}
    \caption{Hyper-parameter sensitivity to confidence threshold $\tau$. We show the evolution of (a) the test accuracy in the target domain w.r.t different setting of $\tau$, (b) the number of pseudo-labels involved in samples from $\mathcal{D}_{u}$ with maximum class probability prediction larger than $\tau$, (c) the number of correct pseudo-labels, and (d) the correction accuracy of pseudo-labels by comparing (b) with (c), while varying the confidence threshold $\tau$. Various colors denote different values with respect to $\tau$. We carry out these experiments on \texttt{DomainNet} in the adaptation scenario ``$R \rightarrow S$'' under the 3-shot setting using ResNet-34.}
    \label{Figure-Sensitivity-Tau}
\end{figure*}

\noindent
{\bf Hyper-parameter Sensitivity to Confidence Threshold $\tau$.} $\tau$ is a hyper-parameter that determines the assignment of pseudo-labels to unlabeled target samples. We study the impact of this parameter on the overall performance of our model. As shown in Figure~\ref{Figure-Sensitivity-Tau}(a)-(d), more pseudo-labels are assigned as $\tau$ decreases, but we can observe an accuracy drop of pseudo-labels at the same time, indicating the adverse effect of the noisy pseudo-labels on the model training. We set $\tau=0.95$ according to its best performance on the validation set. 
 
\begin{figure}[t]
    \centering
    \begin{subfigure}[b]{0.48\textwidth}
        \includegraphics[width=5.71cm,height=2.5cm]{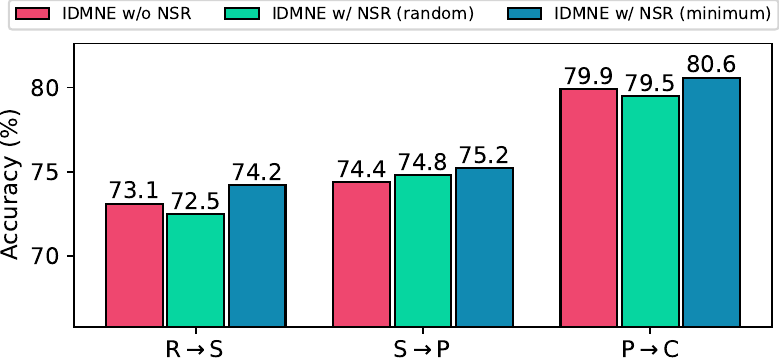}
        \caption*{(a)}
    \end{subfigure}
    \hfill
    \begin{subfigure}[b]{0.24\textwidth}
        \includegraphics[height=2.5cm,width=2.85cm]{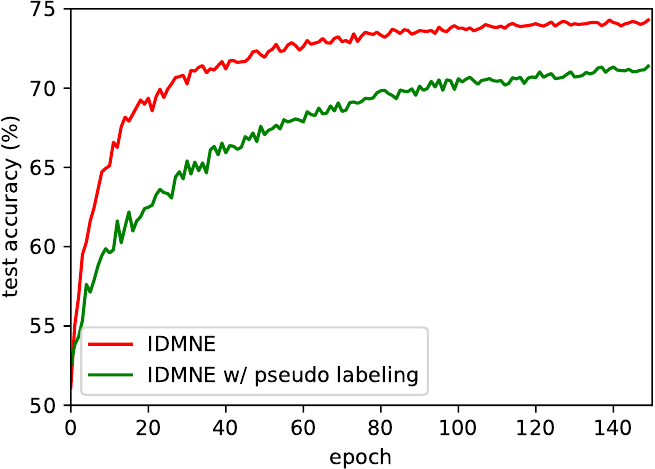}
        \caption*{(b)}
    \end{subfigure}
    \hfill
    \begin{subfigure}[b]{0.24\textwidth}
        \includegraphics[height=2.5cm,width=2.85cm]{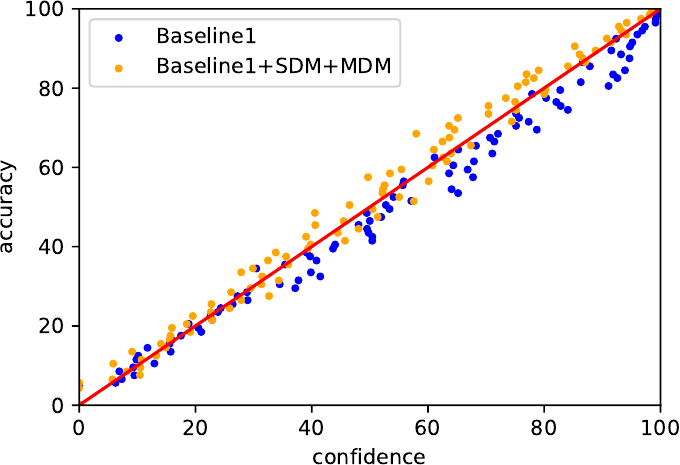}
        \caption*{(c)}
    \end{subfigure}
    \caption{\textcolor{\mycolor}{(a) Comparison results of ``IDMNE w/o NSR'', ``IDMNE w/ NSR (random)'' and ``IDMNE w/ NSR (minimum)''. (b) Comparison results of IDMNE and ``IDMNE w/o pseudo labeling''. (c) Calibration results of ``Baseline1+SDM+MDM'' (with Mixup) and ``Baseline1'' (without Mixup) while displaying with a scatterplot between the accuracy and the average confidence per bin (100 bins in total). The red line indicates where the accuracy matches the confidence. We conducted experiments on \texttt{DomainNet}, specifically in the adaptation cases ``$R \rightarrow S$'', ``$S \rightarrow P$, and ``$P \rightarrow C$'' for (a), as well as in the scenario ``$R \rightarrow S$'' for both (b) and (c). All experiments were performed under the 3-shot setting using the ResNet-34 architecture. (\textbf{Best viewed zoomed in.})}}
    \label{Figure-Ablation-NSR-PL-Calibration}
\end{figure}
 
\noindent
\textcolor{\mycolor}{
\textbf{Effect of the Proposed Complementary Label Selection Scheme in NSR.} 
To evaluate the effectiveness of our proposed complementary label selection scheme, we compared it to the scheme outlined in~\cite{kim2019nlnl}. In their strategy, the complementary label is randomly chosen from all possible class labels, except the one with the highest confidence. Figure~\ref{Figure-Ablation-NSR-PL-Calibration}(a) shows that their scheme, referred to as ``IDMNE w/ NSR (random)'', does not offer a significant performance advantage over our proposed scheme, here denoted as ``IDMNE w/ NSR (minimum)''. In fact, it may even be less effective than the cases indicated by ``IDMNE w/o NSR''  where \textbf{NSR} is not employed.
This outcome can be attributed to the fact that \textbf{NSR} is only applied to unlabeled target samples with a predicted confidence below a predefined threshold. As depicted in Figure~\ref{Figure-Sensitivity-Tau}, the pseudo-labels assigned to these samples tend to be unreliable and may contain noisy labels. Randomly selecting class labels, excluding the one with the highest confidence score, increases the likelihood of selecting a true label as a complementary label. This elevates the risk of propagating incorrect information during NSR, which has an adverse effect on the training of the adaptation model and diminishes its overall performance.
}
 
\noindent
{\bf Necessity of Pseudo Labeling.} We compare IDMNE with ``IDMNE w/o pseudo labeling'', which refers to that the model is trained without using any supervision from pseudo-labels. As shown in Figure~\ref{Figure-Ablation-NSR-PL-Calibration}(b), as training progresses, the accuracy of IDMNE and ``IDMNE w/o pseudo labeling'' gradually improves. However, the accuracy of IDMNE improves faster. At the end, the accuracy of IDMNE surpasses that of the latter by approximately 3\%. This confirms the crucial necessity of pseudo-labels for our proposed method.

\noindent
{\bf Model Calibration with Mixup.} 
According to \cite{guo2017calibration}, deep neural networks (DNNs) tend to possess poor model calibration, resulting in the model generating high probability (confidence) scores for class labels that are actually incorrect. This overconfidence could give rise to the result that the model accuracy on a sample set is lower than its average predicted confidence score on the same set. Thus, overconfidence would do harm to the pseudo labeling scheme since these model predictions with high confidence are likely to produce many noisy pseudo-labels for selected unlabeled target domain samples, thereby causing negative effects in model optimization during training. A well-calibrated model, therefore, is in need. Several recent work~\cite{MixupCalibration, ImprovingCalibration} has demonstrated that Mixup can play a significant role in improving the calibration of DNNs. Here, we conduct experiments to explore the effect of Mixup on model calibration. For simplicity, we only choose ``Baseline1'' (without Mixup) and ``Baseline1+SMD+MDM''(with Mixup) for comparison, both of which have previously been considered in Section~\ref{Subsection:Analysis}. As shown in Figure~\ref{Figure-Ablation-NSR-PL-Calibration}(c), Mixup makes the ``Baseline1+SMD+MDM'' model better calibrated as more sample points lie in or above the red line where the accuracy matches the confidence. Specifically, on samples at a certain confidence level, the accuracy of ``Baseline1+SMD+MDM'' is higher than or at least on par with that of ``Baseline1''. 
 
\begin{figure}[t]
    \centering
    \begin{subfigure}[b]{0.31\textwidth}
        \includegraphics[width=\textwidth]{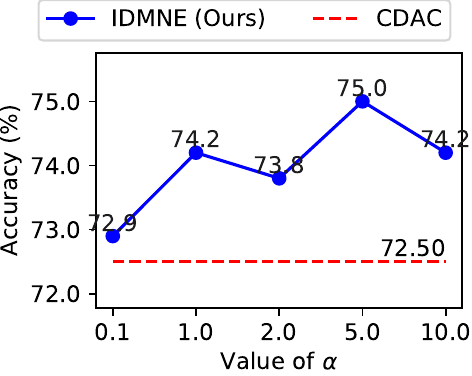}
        \caption*{(a) Sensitivity to $\alpha$}
    \end{subfigure}
    \hfill
    \begin{subfigure}[b]{0.31\textwidth}
        \includegraphics[width=\textwidth]{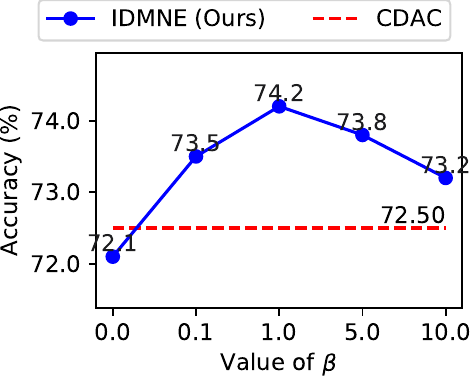}
        \caption*{(b) Sensitivity to $\beta$}
    \end{subfigure}
    \hfill
    \begin{subfigure}[b]{0.31\textwidth}
        \includegraphics[width=\textwidth]{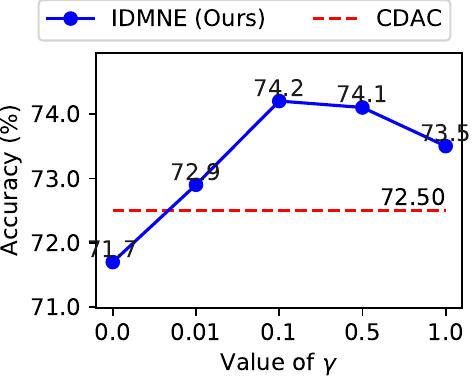}
        \caption*{(c) Sensitivity to $\gamma$}
    \end{subfigure}
    \caption{\textcolor{\mycolor}{Sensitivity to the hyper-parameters $\alpha$, $\beta$, and $\gamma$. The experiments are conducted in the adaptation task of ``$R \rightarrow S$'' on \texttt{DomainNet} with a 3-shot setup using the ResNet-34 backbone.}}
    \label{Figure-Sensitivity}
\end{figure}

\noindent\textcolor{\mycolor}{
\textbf{Hyper-parameter Sensitivity to $\alpha$, $\beta$, and $\gamma$.} We conducted case studies to investigate the sensitivity of the hyper-parameters $\alpha$, $\beta$, and $\gamma$. The results of the sensitivity analysis are displayed in Figure~\ref{Figure-Sensitivity}(a)-(c). We found that in Figure~\ref{Figure-Sensitivity}(a), employing the default value of $\alpha=1.0$, as commonly referenced in MixUp operations~\cite{zhang2018mixup, pmlr-v97-verma19a}, does not yield optimal performance. This indicates that adjusting the value of $\alpha$ can lead to improved results. Notably, even when using the default $\alpha$ value, our method consistently outperforms the state-of-the-art SSDA baseline, \textbf{CDAC}~\cite{li2021cross}, which highlights its robustness to changes in $\alpha$ within this adaptation scenario. Figures~\ref{Figure-Sensitivity}(b)-(c) showcase the impact of the hyper-parameters $\beta$ and $\gamma$ on test accuracy. Initially, increasing the values of $\beta$ and $\gamma$ significantly improves test accuracy. However, further increases gradually diminish accuracy. Nevertheless, the final test accuracy remains superior to that of the \textbf{CDAC} baseline. Overall, our method demonstrates low sensitivity to changes in these two hyper-parameters across a wide range. Notably, the lowest test accuracy is attained when setting both $\beta$ and $\gamma$ to 0. This may result from the excluding of the loss terms associated with Inter-domain Mixup and Neighborhood Expansion from the model training process.}

\begin{figure}[t]
    \centering
    \includegraphics[width=0.75\linewidth]{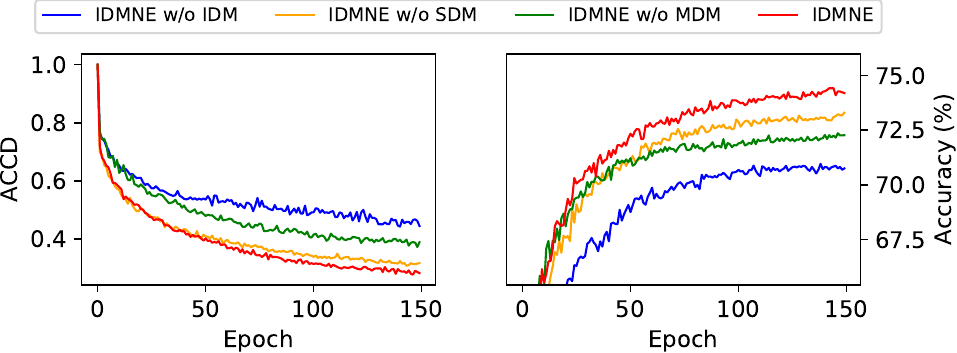}
    \caption{\textcolor{\mycolor}{
    The variations of Averaged Cluster Centroid Distance of ``IDMNE w/o SDM'', ``IDMNE w/o MDM'', and ``IDMNE w/o IDM'' and IDMNE. The experiment is performed on the ``$R \rightarrow S$'' adaptation task of \texttt{DomainNet}, using the ResNet-34 backbone and the 3-shot setup.
}}
    \label{Figure-ACCD}
\end{figure}

\begin{figure}[t]
    \centering
    \includegraphics[width=\linewidth]{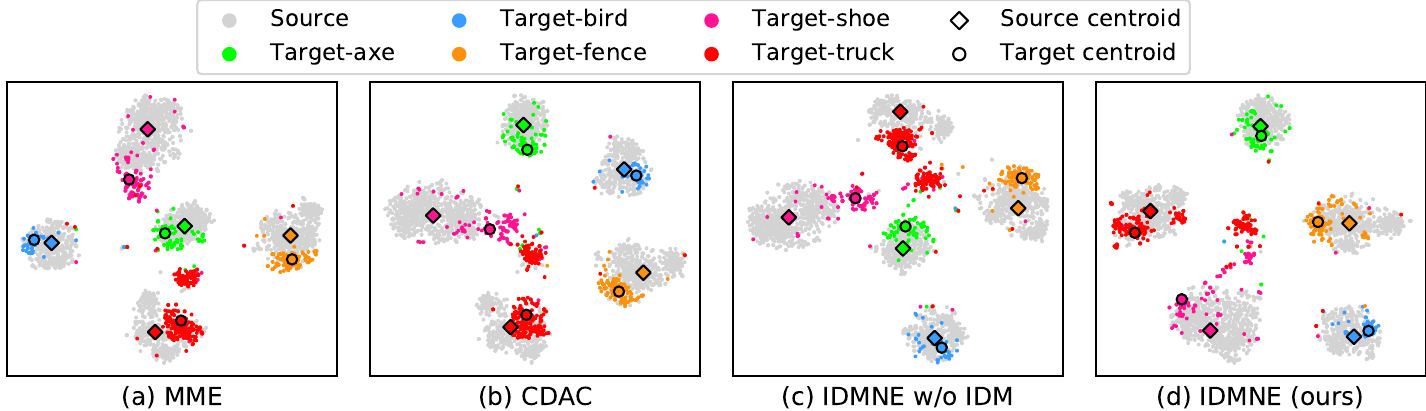}
    \caption{\textcolor{\mycolor}{
    Feature visualization of \textbf{MME}, \textbf{CDAC}, ``IDMNE w/o IDM'' and IDMNE (our full model) using t-SNE. The visualization is performed on the ``$R \rightarrow S$'' adaptation task of \texttt{DomainNet}, using the ResNet-34 backbone and the 3-shot setup. We randomly select five representative classes with distinct bright colors for their demonstration, namely ``Axe'' (green), ``Bir'' (blue), ``Fence'' (orange), ``Shoe'' (pink), and ``Truck'' (red). Additionally, grey data points correspond to source samples, while the cluster centroids of various classes on the source and target domains are represented with ``square'' and ``circle'' markers, respectively.
}}
    \label{Figure-t-SNE}
\end{figure}
 
\noindent
\textcolor{\mycolor}{
\textbf{Feature Distribution and its Visualization.} To gain deeper insights into the effects of each component within Inter-domain Mixup on feature distribution alignment across both domains, we employ visualization techniques and quantitative measures. Specifically, we utilize the Averaged Cluster Centroid Distance (ACCD) method, as elaborated in Section~\ref{Subsection:Protocols}, to quantify the impact of Inter-domain Mixup for domain alignment. Additionally, we conduct feature visualization to provide further support for our analysis. The corresponding results are presented in Figures~\textcolor{red}{\ref{Figure-ACCD}} and~\textcolor{red}{\ref{Figure-t-SNE}}.
}

\textcolor{\mycolor}{
Firstly, ACCD measures the distance between feature clusters of the source and target domains for all classes in the dataset, where smaller ACCDs indicate greater alignment of feature clusters between source and target domains. To assess this, we compared our full model, IDMNE, with three variants: ``IDMNE w/o SDM'', ``IDMNE w/o MDM'', and ``IDMNE w/o ID''. These variants represent degraded versions of IDMNE by removing sample-level data mixing (denoted by \textbf{SDM}), manifold-level data mixing (denoted by \textbf{MDM}), or both (denoted by \textbf{IDM}) from the Inter-domain Mixup.
As shown in Figure~\textcolor{red}{\ref{Figure-ACCD}}, all four models consistently exhibited a gradual decrease in ACCDs during model training, indicating source and target clusters across all classes become closer in the feature space. Notably, ``IDMNE w/o SDM'', ``IDMNE w/o MDM'', and IDMNE achieved better feature alignment than "IDMNE w/o IDM", with IDMNE reaching the minimum ACCD value. This suggests that the proposed Inter-domain Mixup, containing both \textbf{SDM} and \textbf{MDM}, whether used individually or in combination (\textbf{IDM}), contributes to feature alignment across domains, ultimately ensuring the superior classification performance of IDMNE.
}

\textcolor{\mycolor}{
Furthermore, we performed feature visualization using t-SNE for IDMNE, its variant ``IDMNE w/o IDM'', and two comparison methods, \textbf{MME}~\cite{saito2019semi} and \textbf{CDAC}~\cite{li2021cross}. The results, as illustrated in Figure~\textcolor{red}{\ref{Figure-t-SNE}}, clearly demonstrate that IDMNE achieves more compact and aligned feature distributions for samples across both domains, surpassing the performance of \textbf{MME}, \textbf{CDAC}, and ``IDMNE w/o IDM''. Additionally,  in comparison to  ``IDMNE w/o IDM'',  IDMNE exhibits a closer proximity of feature cluster centroids for both domains, highlighting the effectiveness of Inter-domain Mixup in promoting cross-domain feature fusion and achieving superior fusion performance.
}
 
\section{Conclusion}
\label{sec:conclusion}

In this paper, we propose Inter-domain Mixup with Neighborhood Expansion (IDMNE) for semi-supervised domain adaptation of image recognition. IDMNE consists of a well-designed cross-domain feature alignment strategy called Inter-domain Mixup, and a practical auxiliary scheme called Neighborhood Expansion. 
Specifically, Inter-domain Mixup conducts sample-level and manifold-level data mixing of source-target sample pairs from labeled data. Incorporating augmented training samples and their label information reduces cross-domain distribution discrepancies by facilitating cross-domain feature alignment and alleviating label mismatch simultaneously. On the other hand, Neighborhood Expansion leverages massive high-confidence pseudo-labeled samples to diversify the label information of the target domain. Self-Regularization and Pairwise Approaching are also included in Neighborhood Expansion for reducing the uncertainty of model predictions at unlabeled target domain samples. This enables the model to produce higher probability (confidence) scores for predicted class labels corresponding to those unlabeled samples of the target domain. Finally, Inter-domain Mixup and Neighborhood Expansion are integrated into an adaptation framework, and extensive experiments demonstrate that our proposed method achieves considerable accuracy improvement over existing state-of-the-art algorithms on three commonly used benchmark datasets, i.e., \texttt{DomainNet}, \texttt{Office-Home} and \texttt{Office-31}.
 
\bibliography{mybibfile}
 
\end{document}